\pdfoutput=1

\documentclass[11pt]{article}

\usepackage[]{acl}

\usepackage{soul}

\usepackage{times}
\usepackage{latexsym}

\usepackage[T1]{fontenc}

\usepackage[utf8]{inputenc}

\usepackage{microtype}

\usepackage{microtype}

\usepackage[inline]{enumitem}
\usepackage{graphicx}
\usepackage{multirow, booktabs}
\usepackage{soul}
\usepackage{xspace}

\newcommand{\TRUEsize}[0]{11\xspace}
\newcommand{\NumMetrics}[0]{12\xspace}
\newcommand{\wow}[0]{{\sc WoW}}
\newcommand{\summac}{\textsc{SummaC}} 
\newcommand{\summacconv}{SC\textsubscript{Conv}}
\newcommand{\summaczs}{SC\textsubscript{ZS}}

\usepackage{graphics}
\usepackage{amsmath}
\usepackage{xr}

\makeatletter
\newcommand*{\addFileDependency}[1]{
  \typeout{(#1)}
  \@addtofilelist{#1}
  \IfFileExists{#1}{}{\typeout{No file #1.}}
}
\makeatother

\newcommand*{\myexternaldocument}[1]{%
    \externaldocument{#1}%
    \addFileDependency{#1.tex}%
    \addFileDependency{#1.aux}%
}

\myexternaldocument{appendix}

\title{TRUE: Re-evaluating Factual Consistency Evaluation}

\author{
Or Honovich$^{T}$\thanks{\hspace{5px}Work done during an internship at Google Research.} \quad
Roee Aharoni$^{G}$ \quad
\textbf{Jonathan Herzig}$^{G}$ \quad
\textbf{Hagai Taitelbaum}$^{G}$ \\
\textbf{Vered Cohen}$^{G}$ \quad
\textbf{Doron Kukliansky}$^{G}$ \quad
\textbf{Thomas Scialom}$^{M}$ \quad
\textbf{Idan Szpektor}$^{G}$ \\
\textbf{Avinatan Hassidim}$^{G}$ \quad
\textbf{Yossi Matias}$^{G}$ \quad \\
$^T$Tel Aviv University \quad
$^G$Google Research
\quad$^M$Meta AI\\
\quad\quad\quad\quad\quad\quad\quad Tel Aviv, Israel\quad\quad\quad\quad\quad\quad Paris, France\\
{\tt or.honovich@gmail.com} \\ {\tt \{roeeaharoni,szpektor\}@google.com}
}

\begin{document}

\maketitle
\begin{abstract}

Grounded text generation systems often generate text that contains factual inconsistencies, hindering their real-world applicability. Automatic factual consistency evaluation may help alleviate this limitation by accelerating evaluation cycles, filtering inconsistent outputs and augmenting training data. While attracting increasing attention, such evaluation metrics are usually developed and evaluated in silo for a single task or dataset, slowing their adoption. Moreover, previous meta-evaluation protocols focused on system-level correlations with human annotations, which leave the example-level accuracy of such metrics unclear.
In this work, we introduce TRUE: a comprehensive survey and assessment of factual consistency metrics on a standardized collection of existing texts from diverse tasks, manually annotated for factual consistency. Our standardization enables an example-level meta-evaluation protocol that is more actionable and interpretable than previously reported correlations, yielding clearer quality measures. Across diverse state-of-the-art metrics and \TRUEsize{} datasets we find that large-scale NLI and question generation-and-answering-based approaches achieve strong and complementary results. We recommend those methods as a starting point for model and metric developers, and hope TRUE will foster progress towards even better evaluation methods.\footnote{Our code is publicly available at \url{http://www.github.com/google-research/true}}

\end{abstract}

\section{Introduction}\label{sec:introduction}

A core issue in deploying text generation models for real-world applications is that they often generate factually inconsistent text with respect to the input they are conditioned on, or even completely ``hallucinate'' \citep{lee2018hallucinations, rohrbach2018object, maynez-etal-2020-faithfulness, zhao2020reducing} as exemplified in Table~\ref{tab:examples}. 


\begin{table}[t!]
\begin{small}
\begin{center}
\scalebox{0.95}{
\begin{tabular}{ll}
\multicolumn{2}{c}{\textbf{Summarization \cite{wang-etal-2020-asking}}}                                                                                                                                                      \\ \hline
\multicolumn{1}{|l|}{Input}      & \multicolumn{1}{l|}{\begin{tabular}[c]{@{}l@{}}Phyllis schlafly, a leading figure in the \\ us conservative movement, has died at \\ her home in missouri, \textcolor{blue}{aged 92}...\end{tabular}}                      \\ \hline
\multicolumn{1}{|l|}{Summary}    & \multicolumn{1}{l|}{\begin{tabular}[c]{@{}l@{}}Us conservative activist phyllis schlafly \\ has died at the \textcolor{red}{age of 87}.\end{tabular}}                                                                     \\ \hline
                                 &                                                                                                                                                                                                          \\
\multicolumn{2}{c}{\textbf{Fact Verification \cite{thorne-etal-2018-fact}}}                                                                                                                                                                                                           \\ \hline
\multicolumn{1}{|l|}{Evidence}   & \multicolumn{1}{l|}{\begin{tabular}[c]{@{}l@{}}Ronald Bilius ``Ron'' Weasley \textcolor{blue}{is a} \\ \textcolor{blue}{character} in J. K. Rowling's Harry \\ Potter fictional series.\end{tabular}}                                        \\ \hline
\multicolumn{1}{|l|}{Claim}      & \multicolumn{1}{l|}{Ron Weasley \textcolor{red}{is a President}.}                                                                                                                                                         \\ \hline
                                 &                                                                                                                                                                                                          \\
\multicolumn{2}{c}{\textbf{Paraphrasing \cite{zhang-etal-2019-paws}}}                                                                                                                                                                                                            \\ \hline
\multicolumn{1}{|l|}{Input}      & \multicolumn{1}{l|}{\begin{tabular}[c]{@{}l@{}}The tracks were \textcolor{blue}{produced by Tommy} \\ \textcolor{blue}{Lee},  and feature \textcolor{blue}{Michael Beinhorn} \\ \textcolor{blue}{on drums}.\end{tabular}}                                                        \\ \hline
\multicolumn{1}{|l|}{Paraphrase} & \multicolumn{1}{l|}{\begin{tabular}[c]{@{}l@{}}The tracks were \textcolor{red}{produced by Michael} \\ \textcolor{red}{Beinhorn} and have \textcolor{red}{Tommy Lee on} \\ \textcolor{red}{drums}.\end{tabular}}                                                             \\ \hline
                                 &                                                                                                                                                                                                          \\
\multicolumn{2}{c}{\textbf{Knowledge-Grounded Dialogue \cite{honovich-etal-2021-q2}}}                                                                                                                                                                                                                  \\ \hline
\multicolumn{1}{|l|}{Knowledge}  & \multicolumn{1}{l|}{\begin{tabular}[c]{@{}l@{}}The first flip trick called a kickflip, \\ originally called a "magic flip," was \\ \textcolor{blue}{invented by professional skateboarder} \\ \textcolor{blue}{Rodney Mullen}.\end{tabular}} \\ \hline
\multicolumn{1}{|l|}{Response}   & \multicolumn{1}{l|}{\begin{tabular}[c]{@{}l@{}}I remember the first one was called \\ magic flip. It was called a magic flip \\ and was \textcolor{red}{invented in the 60's}.\end{tabular}}                             \\ \hline
\end{tabular}
} 
\end{center}
\end{small}
\caption{Factual inconsistencies (in red) from various tasks which are part of the TRUE study. The corresponding parts in the input/grounding are in blue.}
\label{tab:examples}
\vspace{-0.5cm}
\end{table}

To tackle such inconsistencies, one would like to detect them automatically by predicting whether a generated text is factually consistent with respect to a grounding text (frequently referred to as the ``input'', or the ``knowledge''). 
Such automatic methods attract increasing attention  \citep{zhou-etal-2021-detecting,deng-etal-2021-compression} as they enable both better evaluation and better generation models by automatically filtering training data \citep{gehrmann-etal-2021-gem} or by augmenting training data for controlled generation \citep{rashkin-etal-2021-increasing}.


While automatically evaluating factual consistency is an active line of work, there is no single agreed-upon meta-evaluation protocol for measuring the quality of such methods, and labeling schemes vary in their granularity. Works are usually done in silo, introducing new datasets and methods that target a specific task or domain, such as summarization \citep{falke-etal-2019-ranking,kryscinski-etal-2020-evaluating, wang-etal-2020-asking,scialom-etal-2021-questeval,deutsch2021towards,xie-etal-2021-factual-consistency} or dialogue \citep{dziri2021evaluating, honovich-etal-2021-q2, nie-etal-2021-like, qin-etal-2021-dont}. Comparing the robustness of such methods \textit{across} tasks and datasets is therefore difficult, impeding progress on this subject.



In this work, we present TRUE: a comprehensive survey and assessment of factual consistency evaluation methods, covering various metrics, tasks and datasets. We consolidate \TRUEsize existing datasets annotated for factual consistency into a unified format, including pairs of a target text and a grounding source text, with a binary annotation of whether the target text is factually consistent w.r.t its source. TRUE\footnote{The name is a homage to GLUE \cite{wang-etal-2018-glue} and not an acronym.} covers summarization, knowledge-grounded dialogue, paraphrasing and fact verification.\footnote{We focus on English text-to-text tasks, and leave data-to-text \cite{parikh-etal-2020-totto, reiter-thomson-2020-shared}, multilingual and multimodal tasks to future work.} The proposed standardization enables us to properly compare consistency evaluation methods in a robust manner across these various tasks and domains.

Previous works on automatic factual consistency evaluation have mainly focused on measuring system-level correlations of the proposed metrics with human judgements \cite{pagnoni-etal-2021-understanding}. Yet, these correlations are not useful for estimating the performance of a measured metric when making \textit{example-level, binary} decisions, decoupled from specific system implementations (see recent discussion by \citet{deutsch2022re} on the limitations of reporting correlations).
Instead, we aim to measure how well a method detects inconsistent texts (\textit{recall}) and how often it falsely disregards consistent texts (\textit{precision}), which can be easily computed using the aforementioned binary labeling scheme. 
Therefore, as a meta-evaluation protocol we report the Area Under the ROC Curve (ROC AUC) with respect to inconsistent example detection for each evaluation metric and dataset.

Our thorough survey and assessment of \NumMetrics{} metrics draws a clearer picture on the state of evaluating factual consistency. We show that Natural Language Inference (NLI) approaches, as well as Question Generation and Answering (QG-QA) approaches achieve significantly better\footnote{We conduct significance testing, see section \ref{sec:experiments}.} results on a wide variety of tasks and datasets. We also show that NLI and QG-QA are complementary: combining the two yields even better results and hints at room for further improvement. Finally, we perform both quantitative and qualitative analysis of our results, finding that all approaches struggle with long inputs, labeling issues and personal statements -- paving interesting avenues for future work.



To summarize, our contributions are as follows: 
\begin{enumerate*}[label=(\arabic*)]
  \item  We argue that work on factual consistency evaluation should be unified and generalized across tasks, and standardize \TRUEsize published datasets into a single labeling scheme to corroborate this.
  \item We propose a meta-evaluation protocol that
  allows more actionable and interpretable quality measures than previously reported correlations.
  \item We survey and evaluate \NumMetrics{} diverse metrics in this unified perspective, showing that large-scale NLI and QG-QA-based approaches achieve strong and complementary results \textit{across} tasks.
  \item We analyze our results both qualitatively and quantitatively, pointing at challenges like long inputs and personal statements to be addressed in future work.
\end{enumerate*}

\section{Standardizing Factual Consistency}\label{sec:datasets}

In this section we elaborate on our re-evaluation setup. We first formally define what factual consistency refers to in this work. We then detail the datasets we consider and how we standardize them. Finally, we discuss the meta-evaluation protocol we propose for measuring the performance of evaluation methods on the standardized datasets.

\subsection{Definitions and Terminology}
We define a text to be factually consistent w.r.t its grounding text if all the factual information it conveys is consistent with the factual information conveyed by the grounding text.\footnote{We exclude personal and social statements such as opinions and chit-chat from the scope of factual information.} While some previous works distinguished between inconsistent erroneous text to inconsistent correct text \citep{maynez-etal-2020-faithfulness}, we take a strict approach, requiring the text to be faithful to its grounding text, regardless of the ``correctness'' w.r.t the ``real world''. In other words, we consider only the information present in the input text, not external knowledge, to assess faithfulness. This enables a more well-defined task, since determining the truthfulness of a fact w.r.t a general ``real world'' is subjective and depends on the knowledge, values and beliefs of the subject \cite{heidegger2001essence}. This definition follows similar strictness in Textual Entailment, Question Answering, Summarization and other tasks where comprehension is based on a given grounding text, irrespective of contradiction with other world knowledge. This is also in line with recent work on evaluating attribution in text generation \cite{rashkin2021measuring}, where humans are required to judge whether a generated text is attributable to a grounding text. We use the terms \textit{consistent}, \textit{grounded}, \textit{faithful} and \textit{factual} interchangeably.

\begin{table}[t!]
\begin{center}
\begin{small}
\scalebox{0.73}{
\begin{tabular}{|l|l|l|l|l|l|}
\hline
\textbf{Task}                                 & \textbf{\# Examples}  & \textbf{Open Test} & \textbf{Cons.}    \\ \hline
\textbf{Summarization}                         &                       &                    &                  \\
- FRANK \citep{pagnoni-etal-2021-understanding}   & 671                 & +                 & 33.2\%              \\ 
- SummEval \citep{fabbri2021summeval}             & 1,600               & -                 & 81.6\%            \\
- MNBM \citep{maynez-etal-2020-faithfulness}      & 2,500                 & -                 & 10.2\%              \\
- QAGS-CNNDM \citep{wang-etal-2020-asking}        & 235                   & -                 & 48.1\%             \\ 
- QAGS-XSum \citep{wang-etal-2020-asking}         & 239                   & -                 & 48.5\%              \\ 
\hline
\textbf{Dialogue}                                        &                       &                    &                  \\
- BEGIN \citep{dziri2021evaluating}               & 836                   & +                 & 33.7\%              \\ 
- $Q^2$ \citep{honovich-etal-2021-q2}                    & 1,088                 & -                 & 57.7\%              \\ 
- DialFact \citep{gupta2021dialfact}              & 8,689                 & +                 & 38.5\%            \\ 
\hline 
\textbf{Fact Verification}                               &                       &                    &                  \\
- FEVER \citep{thorne-etal-2018-fact}             & 18,209                & -                 & 35.1\%            \\
- VitaminC \citep{schuster-etal-2021-get}         & 63,054                & +                 & 49.9\%           \\ 
\hline
\textbf{Paraphrasing}                                    &                       &                    &                  \\
- PAWS \citep{zhang-etal-2019-paws}               & 8,000                 & +                 & 44.2\%            \\ 
\hline
\end{tabular}
}
\end{small}
\end{center}
\vspace{-0.25cm}
\caption{Statistics for the datasets incorporated in TRUE. Cons. is the ratio of consistent examples.}
\label{tab:datasets}
\vspace{-0.5cm}
\end{table}

\subsection{Standardization Process}

We include \TRUEsize datasets that contain human annotations w.r.t factual consistency in diverse tasks (Table \ref{tab:datasets}). Other than the importance of covering a wide variety of error types, this also alleviates issues of rating quality which may vary across datasets \cite{denton2021whose}.

To allow a unified evaluation framework we convert all annotations to binary labels that correspond to whether the entire target text is factually consistent w.r.t the given grounding text or not. We note that a fine-grained annotation scheme, i.e., a typology of errors, was proposed for factual consistency \cite{pagnoni-etal-2021-understanding}. While useful, most existing datasets do not include such labels. Moreover, while Machine Translation (MT) evaluation also showed value in fine-grained annotations \cite{freitag2021experts}, it was proposed after years of improving MT to the level where coarse-grained annotation is insufficient. We argue that current grounded generation models are still at early stages w.r.t factual consistency, making binary labeling more beneficial now as it enables easier standardization across tasks and domains, with the goal of bringing researchers to collaborate on a shared methodology. Binary annotation also corresponds to practical applications where filtering out unfaithful predictions is desired, and is in-line with the recommendations for human evaluation of attribution in text generation by \citet{rashkin2021measuring}.

We next detail the \TRUEsize datasets included in TRUE.

\subsubsection{Abstractive Summarization}

\paragraph{FRANK} \citet{pagnoni-etal-2021-understanding} proposed a typology of factual errors, grounded in frame semantics \citep{Fillmore1976FRAMESA, propbank} and linguistic discourse theory \citep{Brown1983BrownGA}. Based on this typology, they collected annotations for model-generated summaries on the CNN/DailyMail \citep[CNN/DM;][]{NIPS2015_cnndm} and XSum \citep{narayan-etal-2018-dont} datasets, resulting in 2250 annotated system outputs. Each summary sentence was annotated by three annotators. We take the majority vote for each sentence to get a sentence-level label and consider a summary as consistent if all sentences are consistent.


\paragraph{SummEval} SummEval \citep{fabbri2020summeval} is a comprehensive study of evaluation metrics for text summarization. The authors collected human judgments for 16 model outputs on 100 articles taken from the CNN/DM dataset, using both extractive and abstractive models. Annotators were asked to rate summaries on a Likert scale from 1 to 5, over 4 dimensions: \textit{consistency}, \textit{coherence}, \textit{fluency} and \textit{relevance}. Each summary was scored by 5 crowd-workers and 3 expert annotators. 
We label summaries as consistent only if all the expert annotators gave a \textit{consistency} score of 5.

\paragraph{MNBM} \citet{maynez-etal-2020-faithfulness} annotated system outputs for the XSum dataset \cite{narayan-etal-2018-dont}. They sampled 500 articles and annotated summaries generated by four different systems, as well as the gold summaries. Annotators were asked to assess whether the summary includes hallucinations.
Judgments from three different annotators were collected for each document-summary pair. 
To convert to a binary-label format, we use the binary consistency decision of whether a summary contains no hallucinations, and assign a label by taking the majority vote of the three annotators.


\paragraph{QAGS} \citet{wang-etal-2020-asking} collected judgments of factual consistency on generated summaries for CNN/DM and XSum. Annotators were presented with the summaries one sentence at a time, along with the article, and determined whether each sentence is factually consistent w.r.t the article. Each sentence was annotated by 3 annotators, using the majority vote as the final score. 
To convert to binary-label format, we consider a summary consistent only if all its sentences are consistent.

\subsubsection{Dialogue Generation}

\paragraph{BEGIN} \citep{dziri2021evaluating} is a dataset for evaluating groundedness in knowledge-grounded dialogue systems, in which system outputs should be consistent with a grounding knowledge provided to the dialogue agent. BEGIN frames the task as textual entailment \citep{dagan-pascal-2003,bowman-etal-2015-large}, adopting the \textit{entailment} and \textit{contradiction} labels, and splitting the neutral label into three sub-categories: \textit{hallucination}, \textit{off-topic} responses and \textit{generic} responses. Dialogue responses were generated by fine-tuning two systems on the {W}izard of {W}ikipedia (\wow{}) dataset \citep{dinan2019wizard}, in which responses should be grounded in a span of text from Wikipedia. The generated responses were split into sentences, and each sentence was annotated separately. 
To convert to a binary-label format, we treat entailed sentences as consistent and all others as inconsistent.

\paragraph{$\boldsymbol{Q^2}$} \citet{honovich-etal-2021-q2} annotated 1,088 generated dialogue responses for binary factual consistency w.r.t the knowledge paragraph provided to the dialogue model, for two dialogue models trained on \wow{}. Responses were annotated using binary labels by 3 of the paper authors, one annotator per response. We use $Q^2$'s labels without changes.

\paragraph{DialFact} \citet{gupta2021dialfact} introduced the task of fact-verification in dialogue and constructed a dataset of conversational claims paired with pieces of evidence from Wikipedia. They define three tasks: (1) detecting whether a response contains verifiable content (2) retrieving relevant evidence and (3) predicting whether a response is \textit{supported} by the evidence, \textit{refuted} by the evidence or if there is \textit{not enough information} to determine. 
We use the verifiable (i.e., factual, rather than personal) responses annotated for the third task, treating \textit{supported} annotations as consistent and the rest as inconsistent. In cases where several evidence were marked as required for verification, we concatenate all evidence sentences to be the grounding text.

\subsubsection{Fact Verification}

\paragraph{FEVER} \citet{thorne-etal-2018-fact} introduced FEVER (Fact Extraction and VERification), a dataset for fact verification against textual sources. FEVER was constructed by extracting information from Wikipedia, generating claims using annotators, then labeling whether each claim is \textit{supported} or \textit{refuted} by Wikipedia. Claims can also be labeled with \textit{NotEnoughInfo}, meaning that there is not enough information in Wikipedia to either verify or refute the claim. Given a claim, the task defined by FEVER is to first extract evidence, then to determine whether it supports or refutes the claim. In a slightly different framing,
the latter stage in FEVER is to determine whether the claim is factually consistent or not w.r.t the evidence, which is aligned with what we aim to measure in TRUE. 
We use the development set of the NLI version of FEVER \citep{nie2019combining, nie-etal-2020-adversarial}, treating \textit{supported} claims as consistent and the rest as inconsistent. 

\paragraph{VitaminC} \citet{schuster-etal-2021-get} derived a large-scale fact verification dataset from factual revisions to Wikipedia pages. Each example includes an evidence text from Wikipedia and a fact, with an annotation of whether the fact is supported, refuted or neutral w.r.t the evidence. The authors collected factual revisions to Wikipedia articles (pairs of ``before'' and ``after'' sentences), and asked annotators to write two facts for each pair: one that is \textit{supported} by the first sentence and \textit{refuted} by the second, and vice versa. When no explicit contradiction was present, the annotators wrote facts that are \textit{neutral} w.r.t the evidence. Additional examples were created by revising examples from FEVER. 
We treat examples that include \textit{supported} facts as consistent, and \textit{refuted} or \textit{neutral} facts as inconsistent.

\subsubsection{Paraphrase Detection}

\paragraph{PAWS} \citet{zhang-etal-2019-paws} constructed a dataset for paraphrase identification with 108,463 paraphrase and non-paraphrase pairs with high lexical overlap, generated by controlled word swapping and back-translation, followed by judgments from human raters. Source sentences were drawn from Wikipedia and the Quora Question Pairs (QQP) corpus. We only use the examples with Wikipedia source sentences and view the binary paraphrase labels as consistency labels. We note that the definition of paraphrase is not equivalent to the definition of factual consistency, as a subset of a source text is not a paraphrase but may still be factually consistent with the source. However, PAWS was constructed such that non-paraphrases usually have contradicting meanings and is therefore relevant.

\subsection{Meta-Evaluation}\label{sec:meta_evaluation}
Previous work on evaluating factual consistency focused on measuring correlation with human judgements \citep{pagnoni-etal-2021-understanding} to compare different metrics. However, such system-level numbers are not very informative when one is interested in evaluating the absolute performance of inconsistency detection methods that perform a \textit{binary} decision w.r.t each input. \citet{deutsch2022re} also recently discuss various issues in measuring system-level correlations to assess the validity of automatic evaluation metrics for summarization.

To conduct a more fine-grained evaluation at the single example level, we report the Receiver Operating Characteristic Area Under the Curve
(ROC AUC) w.r.t binary detection of inconsistent examples.\footnote{This is equivalent to the AUC w.r.t detecting consistent examples.} The ROC curve is created by plotting the \textit{true positive rate} (TPR, a.k.a. the recall) against the \textit{false positive rate} (FPR, a.k.a. the fallout) at different possible thresholds for each tested metric. Measuring ROC AUC evaluates the different metrics without setting a specific decision threshold. 

For datasets with existing development/test split, we also tune a threshold for the binary consistency/inconsistency decision on the development set and report the test set accuracy using this threshold. We tune the thresholds by optimizing the geometric mean of the TPR and 1-FPR: $\sqrt{\text{TPR}*(1-\text{FPR})}$.

\section{Evaluation Metrics}\label{sec:metrics}

We compare various standard as well as state-of-the-art approaches to measure factual consistency. This comparison should draw a clear picture of current research on this subject and raise directions for future work. For example, we expect that robust metrics should perform well across various tasks and datasets. 
We next describe the different metrics we assess as part of this study. We note that for all reference-based metrics, we use the grounding text as the reference. For metrics where the scores are not in the [0,1] range, we normalize the scores to be in that range.


\subsection{N-Gram Based Metrics}
Standard N-Gram matching metrics such as BLEU~\citep{papineni-etal-2002-bleu}, ROUGE~\citep{lin-2004-rouge} and token-level F1 were shown to have weak correlation with factual consistency \cite{maynez-etal-2020-faithfulness, honovich-etal-2021-q2}. We add them as baselines to this study mainly to corroborate this claim on a wide set of datasets and tasks.

\subsection{Model-Based Metrics}

\paragraph{BERTScore} \citep{bert-score} aggregates similarity scores between the BERT contextual embedding of tokens in candidate and reference sentences. We report results for the BERTScore-precision variant as it showed better results in preliminary experiments. We use BERTScore version 0.3.11 with the DeBERTa-xl-MNLI model \citep{he2021deberta, nangia-etal-2017-repeval}, which is the recommended model as of the time of writing this paper.\footnote{\url{https://github.com/Tiiiger/bert_score}}

\paragraph{BLEURT} \cite{sellam-etal-2020-bleurt, sellam-etal-2020-learning} is a learned metric based on BERT \cite{devlin-etal-2019-bert} for evaluating text generation. BLEURT includes additional pretraining on synthetic data followed by fine-tuning on human judgements to train a model that scores system outputs. 
We use the recommended BLEURT-20 checkpoint \citep{pu-etal-2021-learning}.\footnote{\url{https://github.com/google-research/bleurt/blob/master/checkpoints.md}}


\paragraph{FactCC} \citep{kryscinski-etal-2020-evaluating} is a BERT-based metric for verifying the factual consistency of summaries. It is trained on synthetically generated data obtained by applying rule-based transformations to generate consistent and inconsistent summaries.

\paragraph{BARTScore} \citep{yuan2021bartscore} evaluates text using probabilities from force-decoding with a BART model \cite{lewis-etal-2020-bart}. We use the version fine-tuned on the ParaBank2 dataset \citep{hu-etal-2019-large}.

\paragraph{CTC} \citep{deng-etal-2021-compression} measures the average token-level alignment of the generated text w.r.t the grounding text using a BERT sequence tagging model. The model is trained to detect hallucinated tokens generated by a BART model in a self-supervised manner. \footnote{This metric has come to our attention after the paper was accepted, so we add the results to the appendix to avoid adding unreviewed results in the camera ready version. See \url{https://github.com/tanyuqian/ctc-gen-eval/blob/master/factual-consistency.md} for implementation details and Table \ref{tab:auc_weak_metrics} for results.}

\subsection{Natural Language Inference Metrics} 

\paragraph{ANLI} The task of Textual Entailment \cite{dagan-pascal-2003} or Natural Language Inference \citep[NLI;][]{bowman-etal-2015-large} is to determine, given two sentences, a \textit{hypothesis} and a \textit{premise}, whether the \textit{hypothesis} in entailed by the \textit{premise}, contradicts it, or is neutral w.r.t to it. The resemblance\footnote{One example where the definition of NLI and factual consistency differs is in dialog, where subjective or opinionated statements of the dialog agent are not evaluated as factual statements, while NLI models consider all the text in the premise.} of NLI to factual consistency evaluation
has led to utilizing NLI models for measuring factual consistency \citep{thorne-etal-2018-fact, welleck-etal-2019-dialogue, maynez-etal-2020-faithfulness, dziri2021evaluating}. We trained an NLI model by fine-tuning T5-11B \citep{2020t5} on the Adversarial NLI \citep[ANLI;][]{nie-etal-2020-adversarial} dataset. As suggested by \citet{maynez-etal-2020-faithfulness}, we compute the entailment probability with the grounding text as the premise and the generated text as the hypothesis and use it as the example-level factual consistency score.\footnote{More implementation details on the NLI model are available in Section \ref{sec:implementation_details} in the appendix.}

\paragraph{\summac} \citep[Summary Consistency;][]{laban2021summac} is focused on evaluating factual consistency in summarization. They use NLI for detecting inconsistencies by splitting the document and summary into sentences and computing the entailment probabilities on all document/summary sentence pairs, where the premise is a document sentence and the hypothesis is a summary sentence. They aggregate the NLI scores for all pairs by either taking the maximum score per summary sentence and averaging (\summaczs{}) or by training a convolutional neural network to aggregate the scores (\summacconv{}). We use the publicly available implementation.\footnote{\url{https://github.com/tingofurro/summac}}

\subsection{QG-QA Based Metrics} 

\citet{durmus-etal-2020-feqa} and \citet{wang-etal-2020-asking} proposed to use Question Generation (QG) and Question Answering (QA) models to automatically evaluate factual consistency in abstractive summarization, showing promising results. \citet{honovich-etal-2021-q2} employed a similar approach for evaluating knowledge-grounded dialogue generation. 

The steps of the QG-QA approach are as follows: \begin{enumerate*}[label=(\arabic*)] 
\item Questions are automatically generated for spans in the generated text, such that the answer to a question is its respective input span. 
\item The generated questions are answered using a QA model on the grounding text, resulting in an answer span or a ``no-answer'' output.
\item For each question, the two answer spans from the grounding and the generated text are compared to get a score. 
\item The scores for all questions are aggregated into a final score.
\end{enumerate*}

\paragraph{$\mathbf{Q^2}$} \citep{honovich-etal-2021-q2} is a QG-QA method that employs an NLI model to compare the two answers for each question, where the grounding text answer is the premise and the generated text answer is the hypothesis. We report results for a re-implementation of $Q^2$ using T5-11B as the backbone for the QG, QA and NLI models. While \citet{honovich-etal-2021-q2} validate each generated question by answering it using a QA model and comparing to the original extracted answer candidate using exact match, we relax this and instead use F1 token-overlap with a predefined threshold.\footnote{More implementation details are available in Section \ref{sec:implementation_details} in the appendix.}

\paragraph{QuestEval} \cite{scialom-etal-2021-questeval} is a QG-QA method that measures both factual consistency and relevance (by reversing the roles of the generated and grounding texts). The authors trained a model that weights each generated question according to the relevance of its answer to appear in the generated text. Their results showed high correlation with human judgments in comparison to prior work on the SummEval benchmark \citep{fabbri2021summeval}. We use the publicly available version.\footnote{\url{https://github.com/ThomasScialom/QuestEval}}

\begin{table*}[!ht]
\begin{center}
\begin{small}
\scalebox{0.88}{
\begin{tabular}{|l||l||l|l|l|l|l|l|l|l|l|l|}
\hline
                      & \multicolumn{1}{l||}{\textbf{Ensemble}} & \multicolumn{1}{l|}{$\mathbf{Q^2}_{\textbf{metric}}$} & \multicolumn{1}{l|}{\textbf{ANLI}}  & \multicolumn{1}{l|}{\textbf{\summaczs{}}} & \multicolumn{1}{l|}{\textbf{F1}} & \multicolumn{1}{l|}{\textbf{BLEURT}} & \multicolumn{1}{l|}{\textbf{QuestEval}} &  \multicolumn{1}{l|}{\textbf{FactCC}} & \multicolumn{1}{l|}{\textbf{BART$_{\text{score}}$}} & \multicolumn{1}{l|}{\textbf{BERT$_{\text{score}}$}}
\\ \hline\hline
\textbf{FRANK}   & 91.2 & 87.8              & \textbf{89.4}    & 89.1                       & 76.1          & 82.8 & 84.0    & 76.4 & 86.1 & 84.3          \\ \hline
\textbf{SummEval}& 82.9& 78.8              & 80.5             & \textbf{81.7}              & 61.4          & 66.7 & 70.1  & 75.9 & 73.5 & 77.2           \\ \hline
\textbf{MNBM}    & 76.6 & 68.7              & \textbf{77.9**}  & 71.3                       & 46.2          & 64.5 & 65.3  & 59.4 & 60.9 & 62.8          \\ \hline
\textbf{QAGS-C}  & 87.7 & \textbf{83.5}     & 82.1             & 80.9                       & 63.8          & 71.6 & 64.2  & 76.4 & 80.9 & 69.1          \\ \hline
\textbf{QAGS-X}  & 84.8 & 70.9              & \textbf{83.8}    & 78.1                       & 51.1          & 57.2 & 56.3  & 64.9 & 53.8 & 49.5          \\ \hline
\textbf{BEGIN}   &86.2 & 79.7              & 82.6             & 82.0                       & 86.4            & 86.4 & 84.1  & 64.4 & 86.3 & \textbf{87.9} \\ \hline
$\mathbf{Q^2}_{\textbf{dataset}}$   & 82.8 & \textbf{80.9*}    & 72.7             & 77.4                       & 65.9          & 72.4 & 72.2  & 63.7 & 64.9 & 70.0          \\ \hline
\textbf{DialFact}& 90.4 & \textbf{86.1**}   & 77.7             & 84.1                       & 72.3          & 73.1 & 77.3  & 55.3 & 65.6 & 64.2          \\ \hline
\textbf{PAWS}    & 91.2 & \textbf{89.7**}   & 86.4             & 88.2                       & 51.1          & 68.3 & 69.2  & 64.0   & 77.5 & 77.5          \\ \hline\hline
\textbf{FEVER}   & 94.7 & 88.4              & \textbf{93.2**}  & \st{93.2}                  & 51.8          & 59.5 & 72.6  & 61.9 & 64.1 & 63.3          \\ \hline
\textbf{VitaminC}& 96.1 & 81.4              & \textbf{88.3**}  & \st{97.9}                  & 61.4          & 61.8 & 66.5  & 56.3 & 63.2 & 62.5          \\ \hline\hline
\textbf{Avg. $_{\text{w/o VitC, FEVER}}$}    
                 & 86.0 & 80.7              & \textbf{81.5}    & 81.4                       & 63.8          & 71.4 & 71.4 &  66.7 & 72.2 & 71.4          \\ \hline
\end{tabular}
}
\end{small}
\end{center}
\vspace{-0.25cm}
\caption{ROC AUC results for the different metrics on the TRUE development set. We exclude VitaminC and FEVER from the average calculation as \summaczs{} was trained on VitaminC that includes examples from FEVER. The highest score in each row (excluding the Ensemble) is in bold and the aforementioned SC results are in strikethrough. Statistically significant results are indicated using * and ** for $p < 0.05$ and $p < 0.01$ respectively.}
\label{tab:auc}
\end{table*}

\section{Results}\label{sec:experiments}
We report the ROC AUC\footnote{Multiplied by 100 for better readability.} of various metrics on the standardized datasets in Table \ref{tab:auc}. The ROC curves can be found in Figure \ref{fig:roc_curves} in the appendix. \summaczs{} was trained on VitaminC which includes examples from FEVER, so we exclude those datasets from the average AUC calculation for a more fair comparison. As all metrics operate in a ``zero-shot'' manner on all datasets (except for \summaczs{} on VitaminC and FEVER) and no threshold tuning is required, we report results on the development sets.\footnote{AUC for the test sets and accuracy for the dev and test sets are provided in Tables \ref{tab:auc_test}, \ref{tab:acc_dev} and \ref{tab:accuracy_test} in the appendix.} 

The results show that the NLI-based models (ANLI, \summaczs{}\footnote{We report results for \summaczs{} as it performed better in our experiments. Results for \summacconv{} are available in Table \ref{tab:auc_test} in the appendix.}) outperformed the other approaches on 6 datasets, with average AUC of 81.5 and 81.4 for ANLI and \summaczs{}, respectively. ${Q^2}$ outperformed the other approaches on 4 datasets, with an average AUC of 80.7. The next best method, BARTScore, had lower average AUC of 72.2. All other approaches scored 72 or lower on average across all datasets (excluding FEVER and VitaminC). As expected, the simple token-matching based metrics did not perform well, and for completeness, we report their performance in Table \ref{tab:auc_weak_metrics} in the appendix. We keep the F1 score in Table \ref{tab:auc} for convenient comparison to the other metrics. 

One outlier is BEGIN, which is the only dataset where simple metrics like F1 token overlap achieved scores higher than 80. We measured the average overlap between the grounding and target texts per dataset, and found that BEGIN exhibits a high difference between grounded and ungrounded texts in comparison to other datasets (Table \ref{tab:overlap_stat} in appendix \ref{sec:data_statistic}), which explains this.

We follow \citet{laban2021summac} and perform significance testing through bootstrap resampling \cite{efron1982jackknife}, comparing the best method to the second-best method on each dataset. We perform interval comparison at $p = 0.05$ and $p = 0.01$ and find significantly best results on 6 datasets, 3 achieved by ${Q^2}$ and 3 by the ANLI-based model. 

Given that no single method outperformed the rest on all datasets,
we hypothesize that the NLI and QG-QA based metrics are complementary. We test this by averaging the ${Q^2}$, ANLI and \summaczs{} scores per example\footnote{Pairwise ensembles are reported in the appendix, Table~\ref{tab:auc_weak_metrics}.} (\textit{Ensemble} in Table \ref{tab:auc}). Indeed, averaging the three methods yields better results on most datasets and on average, with an increase of 4.5 in ROC AUC from the best single-metric result.

Our results show that a single metric can do well across all tasks and datasets, with all 3 best metrics scoring higher than 80 on average on the 11 datasets. This corroborates our hypothesis that evaluating factual consistency can be unified, and we hope such unified perspective will be adopted in future work to accelerate progress on the subject.

\section{Analysis}\label{sec:analysis}

\paragraph{Input Length.} As QA and NLI models may struggle with long inputs \citep{kocisky-etal-2018-narrativeqa,pang2021quality,yin-etal-2021-docnli,shaham2022scrolls}, metrics based on them may fail when handling long text. To study the effect of input length on the metrics performance, we unify all datasets\footnote{Excluding VitaminC as it is much larger than other datasets and might therefore distort results. Statistics regarding the grounding and target text lengths per dataset is in Appendix \ref{sec:data_statistic}.} and split examples into 6 bins according to the grounding length.\footnote{We measure length in tokens (before subword splitting) as different metrics use different subword tokenizations.} We focus on the grounding as the target texts are usually short (see Table \ref{tab:generated_length} in Appendix \ref{sec:data_statistic}). We measure AUC of the best 3 metrics according to their overall score for each length bin, sampling 1,000 examples per bin.

\begin{figure}
    \includegraphics[width=0.482\textwidth]{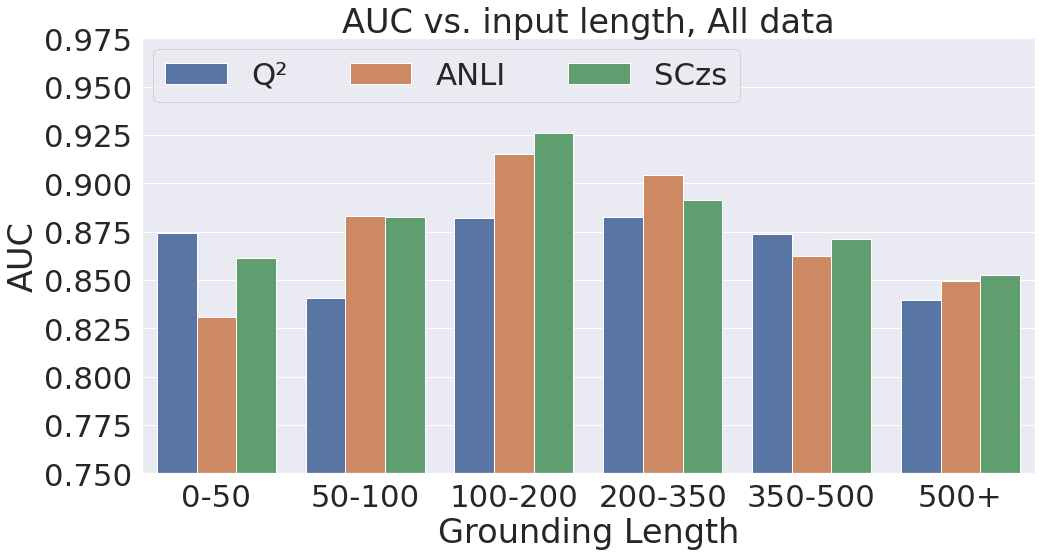}
    \caption{ROC AUC when splitting TRUE's data according to the grounding length.}
    \label{fig:long_input}
\vspace{-0.25cm}
\end{figure}

The results are shown in Figure \ref{fig:long_input}. We find that there is a consistent degradation for texts longer than 200 tokens for all metrics, including \summaczs{} which is designed to better handle long text. We find it surprising that the ANLI-based model and $Q^2$ still do relatively well on the longest bin (with AUC > 0.825) as they perform end-to-end QA and NLI on text with more than 500 tokens.

\paragraph{Model Size.} Model-based metrics are expected to benefit from increasing the model size. To quantify this we study the effect of using smaller models for the ANLI, BLEURT and BERTScore metrics. We compare the average ROC AUC of larger and smaller model variants for each metric. The ablation results are in Table \ref{tab:ablation}. We find an advantage of 4.7, 3.7 and 1.3 average ROC AUC for the larger ANLI, BLEURT and BERTScore variants respectively, showing that larger models indeed allow for better factual consistency evaluation metrics, and hinting at potential improvements from using even larger models.

\begin{table}[ht!]
\begin{center}
\begin{small}
\scalebox{0.9}{
\begin{tabular}{|l|l|}
\hline
\textbf{Model}                   & \textbf{Avg. ROC AUC}          \\ \hline
ANLI-T5-11B                           & \textbf{81.5 (+4.7)}                        \\
ANLI-T5-Large                         & 76.8                        \\ \hline
BLEURT-20                        &  \textbf{71.4 (+3.7)}                        \\ 
BLEURT-20-D6                     & 67.7                        \\ \hline
BERTScore P - deberta-xl-mnli      & \textbf{71.4 (+1.3)}                       \\ 
BERTScore P - roberta-large        & 70.1                        \\ \hline
\end{tabular}
}
\end{small}
\end{center}
\caption{Ablation study comparing the average ROC AUC results for models with different sizes. ``BERTScore P'' stands for BERTScore Precision.}  
\label{tab:ablation}
\vspace{-0.25cm}
\end{table}




\begin{table*}[!ht]
\centering
\resizebox{\textwidth}{!}{
\begin{tabular}{|p{0.45\linewidth}|p{0.4\linewidth}|p{0.45\linewidth}|}
\hline
\textbf{Grounding} & \textbf{Generated Text} & \textbf{Explanation} \\ \hline
The word "philately" is \textcolor{blue}{the English version} of the French word "philatélie", coined by Georges Herpin in 1864. & The word philately is actually \textcolor{red}{a french word} coined by george herpin. &
  The word philately is an English word \textit{based} on a French word, but not French. 
  All best metrics misclassified this. 
  \\ \hline
French police have interviewed presidential candidate francois fillon and his wife penelope over claims \textcolor{blue}{she was paid for fake work.} They provided information that would help find the "truth", mr fillon said… &
French presidential candidate francois fillon has said he and his wife penelope have been questioned by police over claims \textcolor{red}{she worked illegally}. & Most details are correct and the hallucination is subtle. In the case of $Q^2$, most of the generated questions have the same answer based on the grounding and the generated text, therefore the overall score was high. \\ \hline
Stamp collecting is generally accepted as one of the areas that make up the wider subject of \textcolor{blue}{philately, which is the study of stamps.} & 
I've never heard of stamps, but I do know that \textcolor{blue}{the word "philately" refers to the study of stamps.} &
The personal statement “I’ve never heard of stamps”  is not factual and should not be evaluated. \\ \hline
\textcolor{blue}{Evidence suggests that cognitive behavioral therapy and a gradual increase in activity} suited to individual capacity \textcolor{blue}{can be beneficial in some cases.} &
\textcolor{blue}{It has been suggested that cognitive behavioral therapy and gradual increase in exercise could help in some cases} so I'm going to try that for now. &
Similar to the previous examples - SummaC and ANLI falsely marked the text as inconsistent, probably due to the personal statement. \\ \hline
\end{tabular}
}
\caption{Examples for the error analysis. The first two rows show cases of challenging inconsistencies, while the last two show dialogue responses containing non-factual personal statements.}
\label{tab:error_examples}
\end{table*}

\paragraph{Qualitative Analysis.}We conduct manual error analysis to point at weaknesses of the different metrics and present challenges posed by the task.
We analyze 80 examples that were misclassified by all three best metrics, as well as 100 examples that were correctly classified by one or two of the three.

Out of the analyzed examples, many seem to have a wrong label. This is especially true for cases in which all best metrics failed, with annotation errors in 35/80 cases. For the cases where one or two metrics failed, we found annotation errors in 27/100 cases. To verify that the high annotation error rate is indeed a result of inspecting the ``hardest'' examples and not a general issue in the datasets we used, we uniformly sample 100 additional examples, finding that only 10 had annotation errors. We therefore stress that the high misannotation rate indeed characterizes ``hard'' examples only, and is not a general property of the datasets we used. This is inline with the findings of \citet{freitag2021experts}, who showed that in some cases, metrics may be ``better'' than non-expert annotators.
These findings demonstrate the potential of automatic methods in ``cleaning'' training data by filtering factually inconsistent examples.


Despite showing impressive results, the best-performing metrics fail to detect subtle inconsistencies, as presented in Table \ref{tab:error_examples}. This was the case for 21/180 analyzed examples. 
Metrics that aggregate scores across parts of a target text, such as $Q^{2}$ or \summaczs, might assign a high score for texts in which all but a small part is consistent. End-to-end NLI should predict ``contradiction'' even when only a small part of the text contradicts the grounding, but it may fail to do so. Applying a strict approach in the aggregation step, like taking the minimum instead of the average, could potentially remedy this -- with the price of having more false-negatives.
Other errors are caused by domain-specific challenges, such as handling personal statements in dialogues.
As shown in Table \ref{tab:error_examples}, such statements may be falsely classified as ungrounded. This was the case for 10/62 analyzed dialogue responses. A possible way to alleviate this would be to automatically exclude non-factual parts from the evaluation.

\paragraph{Ensemble Analysis.}
As shown in \S\ref{sec:experiments}, a simple averaging ensemble using the three best metrics achieves strong results, outperforming individual metrics on most datasets. To understand this further, we analyze cases in which at least one of the best three metrics failed, while the ensemble succeeded. Overall, there were 25,761 such cases, where in 85.2\% of these cases, two out of the three metrics succeeded, and only one failed. In 14.6\% of these cases, one metric succeeded while the other two failed, and only in 0.2\% of the cases, the ensemble succeeded while all metrics failed. These cases are a result of the different threshold used for the ensemble model vs. the thresholds for the individual metrics.
We sample 100 of these examples and manually analyze them. Out of the sampled examples, 47\% were misclassified by one metric only, where this metric assigned a borderline score - i.e., close to the decision threshold. 36\% of these examples were misclassified by one metric only, and also with a non-borderline score - i.e., the metric was far from a correct prediction. Other cases include two, or even three, erroneous metrics.
\section{Related Work}\label{sec:related}
Adding to the related work mentioned throughout the paper, works on unified evaluation of text generation across tasks include GEM \citep{gehrmann-etal-2021-gem}, where the focus is on evaluating system outputs and not the factual consistency evaluation methods as in TRUE. BEAMetrics \citep{scialom2021beametrics} proposes meta-evaluation protocols across tasks, but does not focus on factual consistency. When discussing consistency (``correctness'') they measure correlations, which are not sufficient as mentioned in Section \ref{sec:meta_evaluation}. \citet{chen-etal-2021-factuality-checkers} present an adversarial meta-evaluation for factual consistency evaluators, focused on summarization.
Other works on meta-evaluation of factual consistency across datasets include GO-FIGURE \citep{gabriel-etal-2021-go} FRANK \citep{pagnoni-etal-2021-understanding} SummaC \citep{laban2021summac} and QAFactEval \citep{fabbri2021qafacteval}, however they all focus solely on summarization. \citet{yeh-etal-2021-comprehensive} conduct a thorough assessment of dialog metrics, however not specifically around factual consistency. To the best of our knowledge, our work is the first to generalize the discussion on evaluating factual consistency across tasks and datasets, and the first to show that large-scale QG-QA and NLI are strong and highly complementary -- setting better baselines and meta-evaluation methodology for future work.

\section{Discussion and Future Work}
We discuss the main takeaways of the TRUE study, pointing at actionable insights for future work. First, as QG-QA and NLI-based methods show better performance than other approaches, especially when combined together, we recommend model developers to use those methods for evaluation when factual consistency is a priority. As for metric developers, we recommend using those methods and the datasets in TRUE when evaluating new metrics. 

We also suggest reporting ROC AUC rather than correlations, as it is more interpretable and actionable. Our proposed binary annotation scheme allows to easily test new metrics across tasks and datasets, which would be useful for future work.

Finally, we encourage data curators to use the binary annotation scheme, which is inline with the recommendations of \citet{rashkin2021measuring}. Having said that, we do not rule out more detailed labeling schemes -- but rather ask to provide a protocol for converting such labels into the more general binary format. 
Future work may also address the challenges of long inputs and personal statements in dialogue, which we point out in our analysis.







\section{Conclusions}
We presented TRUE, a survey and assessment of automatic factual consistency evaluation methods. We standardized various datasets from diverse tasks into a unified labeling scheme to perform a thorough comparison of automatic evaluation methods, showing that large scale NLI and QG-QA based approaches perform well \textit{across} multiple tasks and datasets. We further show that these methods are highly complementary -- hinting at additional headroom for improvement while pointing on current limitations. We hope our results and methodology will encourage a more unified perspective in future work to foster progress towards more factually-consistent NLP applications.

\section*{Acknowledgements}
We thank Dipanjan Das, Sebastian Gehrmann and Joshua Maynez for their valuable comments and suggestions for this work.

\bibliography{anthology,custom}

\begin{thebibliography}{71}
\expandafter\ifx\csname natexlab\endcsname\relax\def\natexlab#1{#1}\fi

\bibitem[{Bowman et~al.(2015)Bowman, Angeli, Potts, and
  Manning}]{bowman-etal-2015-large}
Samuel~R. Bowman, Gabor Angeli, Christopher Potts, and Christopher~D. Manning.
  2015.
\newblock \href {https://doi.org/10.18653/v1/D15-1075} {A large annotated
  corpus for learning natural language inference}.
\newblock In \emph{Proceedings of the 2015 Conference on Empirical Methods in
  Natural Language Processing}, pages 632--642, Lisbon, Portugal. Association
  for Computational Linguistics.

\bibitem[{Brown and Yule(1983)}]{Brown1983BrownGA}
Gillian Brown and George Yule. 1983.
\newblock \emph{Discourse Analysis}.
\newblock Cambridge University Press.

\bibitem[{Chen et~al.(2021)Chen, Liu, and
  Qiu}]{chen-etal-2021-factuality-checkers}
Yiran Chen, Pengfei Liu, and Xipeng Qiu. 2021.
\newblock \href {https://aclanthology.org/2021.findings-emnlp.179} {Are
  factuality checkers reliable? adversarial meta-evaluation of factuality in
  summarization}.
\newblock In \emph{Findings of the Association for Computational Linguistics:
  EMNLP 2021}, pages 2082--2095, Punta Cana, Dominican Republic. Association
  for Computational Linguistics.

\bibitem[{Dagan et~al.(2006)Dagan, Glickman, and Magnini}]{dagan-pascal-2003}
Ido Dagan, Oren Glickman, and Bernardo Magnini. 2006.
\newblock The pascal recognising textual entailment challenge.
\newblock In \emph{Machine Learning Challenges. Evaluating Predictive
  Uncertainty, Visual Object Classification, and Recognising Tectual
  Entailment}, pages 177--190, Berlin, Heidelberg. Springer Berlin Heidelberg.

\bibitem[{Deng et~al.(2021)Deng, Tan, Liu, Xing, and
  Hu}]{deng-etal-2021-compression}
Mingkai Deng, Bowen Tan, Zhengzhong Liu, Eric Xing, and Zhiting Hu. 2021.
\newblock \href {https://aclanthology.org/2021.emnlp-main.599} {Compression,
  transduction, and creation: A unified framework for evaluating natural
  language generation}.
\newblock In \emph{Proceedings of the 2021 Conference on Empirical Methods in
  Natural Language Processing}, pages 7580--7605, Online and Punta Cana,
  Dominican Republic. Association for Computational Linguistics.

\bibitem[{Denton et~al.(2021)Denton, D{\'\i}az, Kivlichan, Prabhakaran, and
  Rosen}]{denton2021whose}
Emily Denton, Mark D{\'\i}az, Ian Kivlichan, Vinodkumar Prabhakaran, and Rachel
  Rosen. 2021.
\newblock Whose ground truth? accounting for individual and collective
  identities underlying dataset annotation.
\newblock \emph{arXiv preprint arXiv:2112.04554}.

\bibitem[{Deutsch et~al.(2021)Deutsch, Bedrax-Weiss, and
  Roth}]{deutsch2021towards}
Daniel Deutsch, Tania Bedrax-Weiss, and Dan Roth. 2021.
\newblock Towards question-answering as an automatic metric for evaluating the
  content quality of a summary.
\newblock \emph{Transactions of the Association for Computational Linguistics},
  9:774--789.

\bibitem[{Deutsch et~al.(2022)Deutsch, Dror, and Roth}]{deutsch2022re}
Daniel Deutsch, Rotem Dror, and Dan Roth. 2022.
\newblock Re-examining system-level correlations of automatic summarization
  evaluation metrics.
\newblock \emph{arXiv preprint arXiv:2204.10216}.

\bibitem[{Devlin et~al.(2019)Devlin, Chang, Lee, and
  Toutanova}]{devlin-etal-2019-bert}
Jacob Devlin, Ming-Wei Chang, Kenton Lee, and Kristina Toutanova. 2019.
\newblock \href {https://doi.org/10.18653/v1/N19-1423} {{BERT}: Pre-training of
  deep bidirectional transformers for language understanding}.
\newblock In \emph{Proceedings of the 2019 Conference of the North {A}merican
  Chapter of the Association for Computational Linguistics: Human Language
  Technologies, Volume 1 (Long and Short Papers)}, pages 4171--4186,
  Minneapolis, Minnesota. Association for Computational Linguistics.

\bibitem[{Dinan et~al.(2019)Dinan, Roller, Shuster, Fan, Auli, and
  Weston}]{dinan2019wizard}
Emily Dinan, Stephen Roller, Kurt Shuster, Angela Fan, Michael Auli, and Jason
  Weston. 2019.
\newblock {W}izard of {W}ikipedia: Knowledge-powered conversational agents.
\newblock In \emph{Proceedings of the International Conference on Learning
  Representations (ICLR)}.

\bibitem[{Durmus et~al.(2020)Durmus, He, and Diab}]{durmus-etal-2020-feqa}
Esin Durmus, He~He, and Mona Diab. 2020.
\newblock \href {https://doi.org/10.18653/v1/2020.acl-main.454} {{FEQA}: A
  question answering evaluation framework for faithfulness assessment in
  abstractive summarization}.
\newblock In \emph{Proceedings of the 58th Annual Meeting of the Association
  for Computational Linguistics}, pages 5055--5070, Online. Association for
  Computational Linguistics.

\bibitem[{Dziri et~al.(2021)Dziri, Rashkin, Linzen, and
  Reitter}]{dziri2021evaluating}
Nouha Dziri, Hannah Rashkin, Tal Linzen, and David Reitter. 2021.
\newblock \href {http://arxiv.org/abs/2105.00071} {Evaluating groundedness in
  dialogue systems: The begin benchmark}.

\bibitem[{Efron(1982)}]{efron1982jackknife}
Bradley Efron. 1982.
\newblock \emph{The jackknife, the bootstrap and other resampling plans}.
\newblock SIAM.

\bibitem[{Fabbri et~al.(2020)Fabbri, Kry{\'s}ci{\'n}ski, McCann, Xiong, Socher,
  and Radev}]{fabbri2020summeval}
Alexander~R Fabbri, Wojciech Kry{\'s}ci{\'n}ski, Bryan McCann, Caiming Xiong,
  Richard Socher, and Dragomir Radev. 2020.
\newblock Summeval: Re-evaluating summarization evaluation.
\newblock \emph{arXiv preprint arXiv:2007.12626}.

\bibitem[{Fabbri et~al.(2021{\natexlab{a}})Fabbri, Kry{\'s}ci{\'n}ski, McCann,
  Xiong, Socher, and Radev}]{fabbri2021summeval}
Alexander~R Fabbri, Wojciech Kry{\'s}ci{\'n}ski, Bryan McCann, Caiming Xiong,
  Richard Socher, and Dragomir Radev. 2021{\natexlab{a}}.
\newblock Summeval: Re-evaluating summarization evaluation.
\newblock \emph{Transactions of the Association for Computational Linguistics},
  9:391--409.

\bibitem[{Fabbri et~al.(2021{\natexlab{b}})Fabbri, Wu, Liu, and
  Xiong}]{fabbri2021qafacteval}
Alexander~R. Fabbri, Chien-Sheng Wu, Wenhao Liu, and Caiming Xiong.
  2021{\natexlab{b}}.
\newblock \href {http://arxiv.org/abs/2112.08542} {Qafacteval: Improved
  qa-based factual consistency evaluation for summarization}.

\bibitem[{Falke et~al.(2019)Falke, Ribeiro, Utama, Dagan, and
  Gurevych}]{falke-etal-2019-ranking}
Tobias Falke, Leonardo F.~R. Ribeiro, Prasetya~Ajie Utama, Ido Dagan, and Iryna
  Gurevych. 2019.
\newblock \href {https://doi.org/10.18653/v1/P19-1213} {Ranking generated
  summaries by correctness: An interesting but challenging application for
  natural language inference}.
\newblock In \emph{Proceedings of the 57th Annual Meeting of the Association
  for Computational Linguistics}, pages 2214--2220, Florence, Italy.
  Association for Computational Linguistics.

\bibitem[{Fillmore(1976)}]{Fillmore1976FRAMESA}
C.~Fillmore. 1976.
\newblock Frame semantics and the nature of language *.
\newblock \emph{Annals of the New York Academy of Sciences}, 280.

\bibitem[{Freitag et~al.(2021)Freitag, Foster, Grangier, Ratnakar, Tan, and
  Macherey}]{freitag2021experts}
Markus Freitag, George Foster, David Grangier, Viresh Ratnakar, Qijun Tan, and
  Wolfgang Macherey. 2021.
\newblock Experts, errors, and context: A large-scale study of human evaluation
  for machine translation.
\newblock \emph{arXiv preprint arXiv:2104.14478}.

\bibitem[{Gabriel et~al.(2021)Gabriel, Celikyilmaz, Jha, Choi, and
  Gao}]{gabriel-etal-2021-go}
Saadia Gabriel, Asli Celikyilmaz, Rahul Jha, Yejin Choi, and Jianfeng Gao.
  2021.
\newblock \href {https://doi.org/10.18653/v1/2021.findings-acl.42} {{GO}
  {FIGURE}: A meta evaluation of factuality in summarization}.
\newblock In \emph{Findings of the Association for Computational Linguistics:
  ACL-IJCNLP 2021}, pages 478--487, Online. Association for Computational
  Linguistics.

\bibitem[{Gehrmann et~al.(2021)Gehrmann, Adewumi, Aggarwal, Ammanamanchi,
  Aremu, Bosselut, Chandu, Clinciu, Das, Dhole, Du, Durmus, Du{\v{s}}ek,
  Emezue, Gangal, Garbacea, Hashimoto, Hou, Jernite, Jhamtani, Ji, Jolly, Kale,
  Kumar, Ladhak, Madaan, Maddela, Mahajan, Mahamood, Majumder, Martins,
  McMillan-Major, Mille, van Miltenburg, Nadeem, Narayan, Nikolaev,
  Niyongabo~Rubungo, Osei, Parikh, Perez-Beltrachini, Rao, Raunak, Rodriguez,
  Santhanam, Sedoc, Sellam, Shaikh, Shimorina, Sobrevilla~Cabezudo, Strobelt,
  Subramani, Xu, Yang, Yerukola, and Zhou}]{gehrmann-etal-2021-gem}
Sebastian Gehrmann, Tosin Adewumi, Karmanya Aggarwal, Pawan~Sasanka
  Ammanamanchi, Anuoluwapo Aremu, Antoine Bosselut, Khyathi~Raghavi Chandu,
  Miruna-Adriana Clinciu, Dipanjan Das, Kaustubh Dhole, Wanyu Du, Esin Durmus,
  Ond{\v{r}}ej Du{\v{s}}ek, Chris~Chinenye Emezue, Varun Gangal, Cristina
  Garbacea, Tatsunori Hashimoto, Yufang Hou, Yacine Jernite, Harsh Jhamtani,
  Yangfeng Ji, Shailza Jolly, Mihir Kale, Dhruv Kumar, Faisal Ladhak, Aman
  Madaan, Mounica Maddela, Khyati Mahajan, Saad Mahamood, Bodhisattwa~Prasad
  Majumder, Pedro~Henrique Martins, Angelina McMillan-Major, Simon Mille, Emiel
  van Miltenburg, Moin Nadeem, Shashi Narayan, Vitaly Nikolaev, Andre
  Niyongabo~Rubungo, Salomey Osei, Ankur Parikh, Laura Perez-Beltrachini,
  Niranjan~Ramesh Rao, Vikas Raunak, Juan~Diego Rodriguez, Sashank Santhanam,
  Jo{\~a}o Sedoc, Thibault Sellam, Samira Shaikh, Anastasia Shimorina,
  Marco~Antonio Sobrevilla~Cabezudo, Hendrik Strobelt, Nishant Subramani, Wei
  Xu, Diyi Yang, Akhila Yerukola, and Jiawei Zhou. 2021.
\newblock \href {https://doi.org/10.18653/v1/2021.gem-1.10} {The {GEM}
  benchmark: Natural language generation, its evaluation and metrics}.
\newblock In \emph{Proceedings of the 1st Workshop on Natural Language
  Generation, Evaluation, and Metrics (GEM 2021)}, pages 96--120, Online.
  Association for Computational Linguistics.

\bibitem[{Gupta et~al.(2021)Gupta, Wu, Liu, and Xiong}]{gupta2021dialfact}
Prakhar Gupta, Chien-Sheng Wu, Wenhao Liu, and Caiming Xiong. 2021.
\newblock Dialfact: A benchmark for fact-checking in dialogue.
\newblock \emph{arXiv preprint arXiv:2110.08222}.

\bibitem[{He et~al.(2021)He, Liu, Gao, and Chen}]{he2021deberta}
Pengcheng He, Xiaodong Liu, Jianfeng Gao, and Weizhu Chen. 2021.
\newblock \href {https://openreview.net/forum?id=XPZIaotutsD} {Deberta:
  Decoding-enhanced bert with disentangled attention}.
\newblock In \emph{International Conference on Learning Representations}.

\bibitem[{Heidegger(2001)}]{heidegger2001essence}
Martin Heidegger. 2001.
\newblock On the essence of truth.
\newblock \emph{The Nature of Truth: Classic and Contemporary Perspectives},
  1:295--316.

\bibitem[{Hermann et~al.(2015)Hermann, Kocisky, Grefenstette, Espeholt, Kay,
  Suleyman, and Blunsom}]{NIPS2015_cnndm}
Karl~Moritz Hermann, Tomas Kocisky, Edward Grefenstette, Lasse Espeholt, Will
  Kay, Mustafa Suleyman, and Phil Blunsom. 2015.
\newblock \href
  {https://proceedings.neurips.cc/paper/2015/file/afdec7005cc9f14302cd0474fd0f3c96-Paper.pdf}
  {Teaching machines to read and comprehend}.
\newblock In \emph{Advances in Neural Information Processing Systems},
  volume~28. Curran Associates, Inc.

\bibitem[{Honovich et~al.(2021)Honovich, Choshen, Aharoni, Neeman, Szpektor,
  and Abend}]{honovich-etal-2021-q2}
Or~Honovich, Leshem Choshen, Roee Aharoni, Ella Neeman, Idan Szpektor, and Omri
  Abend. 2021.
\newblock \href {https://aclanthology.org/2021.emnlp-main.619} {$q^{2}$:
  {E}valuating factual consistency in knowledge-grounded dialogues via question
  generation and question answering}.
\newblock In \emph{Proceedings of the 2021 Conference on Empirical Methods in
  Natural Language Processing}, pages 7856--7870, Online and Punta Cana,
  Dominican Republic. Association for Computational Linguistics.

\bibitem[{Hu et~al.(2019)Hu, Singh, Holzenberger, Post, and
  Van~Durme}]{hu-etal-2019-large}
J.~Edward Hu, Abhinav Singh, Nils Holzenberger, Matt Post, and Benjamin
  Van~Durme. 2019.
\newblock \href {https://doi.org/10.18653/v1/K19-1005} {Large-scale, diverse,
  paraphrastic bitexts via sampling and clustering}.
\newblock In \emph{Proceedings of the 23rd Conference on Computational Natural
  Language Learning (CoNLL)}, pages 44--54, Hong Kong, China. Association for
  Computational Linguistics.

\bibitem[{Ko{\v{c}}isk{\'y} et~al.(2018)Ko{\v{c}}isk{\'y}, Schwarz, Blunsom,
  Dyer, Hermann, Melis, and Grefenstette}]{kocisky-etal-2018-narrativeqa}
Tom{\'a}{\v{s}} Ko{\v{c}}isk{\'y}, Jonathan Schwarz, Phil Blunsom, Chris Dyer,
  Karl~Moritz Hermann, G{\'a}bor Melis, and Edward Grefenstette. 2018.
\newblock \href {https://doi.org/10.1162/tacl_a_00023} {The {N}arrative{QA}
  reading comprehension challenge}.
\newblock \emph{Transactions of the Association for Computational Linguistics},
  6:317--328.

\bibitem[{Kryscinski et~al.(2020)Kryscinski, McCann, Xiong, and
  Socher}]{kryscinski-etal-2020-evaluating}
Wojciech Kryscinski, Bryan McCann, Caiming Xiong, and Richard Socher. 2020.
\newblock \href {https://doi.org/10.18653/v1/2020.emnlp-main.750} {Evaluating
  the factual consistency of abstractive text summarization}.
\newblock In \emph{Proceedings of the 2020 Conference on Empirical Methods in
  Natural Language Processing (EMNLP)}, pages 9332--9346, Online. Association
  for Computational Linguistics.

\bibitem[{Laban et~al.(2021)Laban, Schnabel, Bennett, and
  Hearst}]{laban2021summac}
Philippe Laban, Tobias Schnabel, Paul~N Bennett, and Marti~A Hearst. 2021.
\newblock Summac: Re-visiting nli-based models for inconsistency detection in
  summarization.
\newblock \emph{arXiv preprint arXiv:2111.09525}.

\bibitem[{Lan et~al.(2019)Lan, Chen, Goodman, Gimpel, Sharma, and
  Soricut}]{lan2019albert}
Zhenzhong Lan, Mingda Chen, Sebastian Goodman, Kevin Gimpel, Piyush Sharma, and
  Radu Soricut. 2019.
\newblock Albert: A lite bert for self-supervised learning of language
  representations.
\newblock \emph{arXiv preprint arXiv:1909.11942}.

\bibitem[{Lee et~al.(2018)Lee, Firat, Agarwal, Fannjiang, and
  Sussillo}]{lee2018hallucinations}
Katherine Lee, Orhan Firat, Ashish Agarwal, Clara Fannjiang, and David
  Sussillo. 2018.
\newblock Hallucinations in neural machine translation.
\newblock \emph{NeurIPS 2018 Workshop on Interpretability and Robustness for
  Audio, Speech, and Language.}

\bibitem[{Lewis et~al.(2020)Lewis, Liu, Goyal, Ghazvininejad, Mohamed, Levy,
  Stoyanov, and Zettlemoyer}]{lewis-etal-2020-bart}
Mike Lewis, Yinhan Liu, Naman Goyal, Marjan Ghazvininejad, Abdelrahman Mohamed,
  Omer Levy, Veselin Stoyanov, and Luke Zettlemoyer. 2020.
\newblock \href {https://doi.org/10.18653/v1/2020.acl-main.703} {{BART}:
  Denoising sequence-to-sequence pre-training for natural language generation,
  translation, and comprehension}.
\newblock In \emph{Proceedings of the 58th Annual Meeting of the Association
  for Computational Linguistics}, pages 7871--7880, Online. Association for
  Computational Linguistics.

\bibitem[{Lin(2004)}]{lin-2004-rouge}
Chin-Yew Lin. 2004.
\newblock \href {https://aclanthology.org/W04-1013} {{ROUGE}: A package for
  automatic evaluation of summaries}.
\newblock In \emph{Text Summarization Branches Out}, pages 74--81, Barcelona,
  Spain. Association for Computational Linguistics.

\bibitem[{Liu et~al.(2019)Liu, Ott, Goyal, Du, Joshi, Chen, Levy, Lewis,
  Zettlemoyer, and Stoyanov}]{liu2019roberta}
Yinhan Liu, Myle Ott, Naman Goyal, Jingfei Du, Mandar Joshi, Danqi Chen, Omer
  Levy, Mike Lewis, Luke Zettlemoyer, and Veselin Stoyanov. 2019.
\newblock Roberta: A robustly optimized bert pretraining approach.
\newblock \emph{arXiv preprint arXiv:1907.11692}.

\bibitem[{Maynez et~al.(2020)Maynez, Narayan, Bohnet, and
  McDonald}]{maynez-etal-2020-faithfulness}
Joshua Maynez, Shashi Narayan, Bernd Bohnet, and Ryan McDonald. 2020.
\newblock \href {https://doi.org/10.18653/v1/2020.acl-main.173} {On
  faithfulness and factuality in abstractive summarization}.
\newblock In \emph{Proceedings of the 58th Annual Meeting of the Association
  for Computational Linguistics}, pages 1906--1919, Online. Association for
  Computational Linguistics.

\bibitem[{Nangia et~al.(2017)Nangia, Williams, Lazaridou, and
  Bowman}]{nangia-etal-2017-repeval}
Nikita Nangia, Adina Williams, Angeliki Lazaridou, and Samuel Bowman. 2017.
\newblock \href {https://doi.org/10.18653/v1/W17-5301} {The {R}ep{E}val 2017
  shared task: Multi-genre natural language inference with sentence
  representations}.
\newblock In \emph{Proceedings of the 2nd Workshop on Evaluating Vector Space
  Representations for {NLP}}, pages 1--10, Copenhagen, Denmark. Association for
  Computational Linguistics.

\bibitem[{Narayan et~al.(2018)Narayan, Cohen, and
  Lapata}]{narayan-etal-2018-dont}
Shashi Narayan, Shay~B. Cohen, and Mirella Lapata. 2018.
\newblock \href {https://doi.org/10.18653/v1/D18-1206} {Don{'}t give me the
  details, just the summary! topic-aware convolutional neural networks for
  extreme summarization}.
\newblock In \emph{Proceedings of the 2018 Conference on Empirical Methods in
  Natural Language Processing}, pages 1797--1807, Brussels, Belgium.
  Association for Computational Linguistics.

\bibitem[{Nie et~al.(2019)Nie, Chen, and Bansal}]{nie2019combining}
Yixin Nie, Haonan Chen, and Mohit Bansal. 2019.
\newblock Combining fact extraction and verification with neural semantic
  matching networks.
\newblock In \emph{Association for the Advancement of Artificial Intelligence
  ({AAAI})}.

\bibitem[{Nie et~al.(2020)Nie, Williams, Dinan, Bansal, Weston, and
  Kiela}]{nie-etal-2020-adversarial}
Yixin Nie, Adina Williams, Emily Dinan, Mohit Bansal, Jason Weston, and Douwe
  Kiela. 2020.
\newblock \href {https://doi.org/10.18653/v1/2020.acl-main.441} {Adversarial
  {NLI}: A new benchmark for natural language understanding}.
\newblock In \emph{Proceedings of the 58th Annual Meeting of the Association
  for Computational Linguistics}, pages 4885--4901, Online. Association for
  Computational Linguistics.

\bibitem[{Nie et~al.(2021)Nie, Williamson, Bansal, Kiela, and
  Weston}]{nie-etal-2021-like}
Yixin Nie, Mary Williamson, Mohit Bansal, Douwe Kiela, and Jason Weston. 2021.
\newblock \href {https://doi.org/10.18653/v1/2021.acl-long.134} {{I} like fish,
  especially dolphins: Addressing contradictions in dialogue modeling}.
\newblock In \emph{Proceedings of the 59th Annual Meeting of the Association
  for Computational Linguistics and the 11th International Joint Conference on
  Natural Language Processing (Volume 1: Long Papers)}, pages 1699--1713,
  Online. Association for Computational Linguistics.

\bibitem[{Pagnoni et~al.(2021)Pagnoni, Balachandran, and
  Tsvetkov}]{pagnoni-etal-2021-understanding}
Artidoro Pagnoni, Vidhisha Balachandran, and Yulia Tsvetkov. 2021.
\newblock \href {https://doi.org/10.18653/v1/2021.naacl-main.383}
  {Understanding factuality in abstractive summarization with {FRANK}: A
  benchmark for factuality metrics}.
\newblock In \emph{Proceedings of the 2021 Conference of the North American
  Chapter of the Association for Computational Linguistics: Human Language
  Technologies}, pages 4812--4829, Online. Association for Computational
  Linguistics.

\bibitem[{Palmer et~al.(2005)Palmer, Gildea, and Kingsbury}]{propbank}
Martha Palmer, Daniel Gildea, and Paul Kingsbury. 2005.
\newblock \href {https://doi.org/10.1162/0891201053630264} {{The Proposition
  Bank: An Annotated Corpus of Semantic Roles}}.
\newblock \emph{Computational Linguistics}, 31(1):71--106.

\bibitem[{Pang et~al.(2021)Pang, Parrish, Joshi, Nangia, Phang, Chen,
  Padmakumar, Ma, Thompson, He et~al.}]{pang2021quality}
Richard~Yuanzhe Pang, Alicia Parrish, Nitish Joshi, Nikita Nangia, Jason Phang,
  Angelica Chen, Vishakh Padmakumar, Johnny Ma, Jana Thompson, He~He, et~al.
  2021.
\newblock Quality: Question answering with long input texts, yes!
\newblock \emph{arXiv preprint arXiv:2112.08608}.

\bibitem[{Papineni et~al.(2002)Papineni, Roukos, Ward, and
  Zhu}]{papineni-etal-2002-bleu}
Kishore Papineni, Salim Roukos, Todd Ward, and Wei-Jing Zhu. 2002.
\newblock \href {https://doi.org/10.3115/1073083.1073135} {{B}leu: a method for
  automatic evaluation of machine translation}.
\newblock In \emph{Proceedings of the 40th Annual Meeting of the Association
  for Computational Linguistics}, pages 311--318, Philadelphia, Pennsylvania,
  USA. Association for Computational Linguistics.

\bibitem[{Parikh et~al.(2020)Parikh, Wang, Gehrmann, Faruqui, Dhingra, Yang,
  and Das}]{parikh-etal-2020-totto}
Ankur Parikh, Xuezhi Wang, Sebastian Gehrmann, Manaal Faruqui, Bhuwan Dhingra,
  Diyi Yang, and Dipanjan Das. 2020.
\newblock \href {https://doi.org/10.18653/v1/2020.emnlp-main.89} {{ToTTo}: A
  controlled table-to-text generation dataset}.
\newblock In \emph{Proceedings of the 2020 Conference on Empirical Methods in
  Natural Language Processing (EMNLP)}, pages 1173--1186, Online. Association
  for Computational Linguistics.

\bibitem[{Pu et~al.(2021)Pu, Chung, Parikh, Gehrmann, and
  Sellam}]{pu-etal-2021-learning}
Amy Pu, Hyung~Won Chung, Ankur Parikh, Sebastian Gehrmann, and Thibault Sellam.
  2021.
\newblock \href {https://aclanthology.org/2021.emnlp-main.58} {Learning compact
  metrics for {MT}}.
\newblock In \emph{Proceedings of the 2021 Conference on Empirical Methods in
  Natural Language Processing}, pages 751--762, Online and Punta Cana,
  Dominican Republic. Association for Computational Linguistics.

\bibitem[{Qin et~al.(2021)Qin, Xie, Huang, Chen, Xu, and
  Che}]{qin-etal-2021-dont}
Libo Qin, Tianbao Xie, Shijue Huang, Qiguang Chen, Xiao Xu, and Wanxiang Che.
  2021.
\newblock \href {https://aclanthology.org/2021.emnlp-main.182} {Don{'}t be
  contradicted with anything! {CI}-{T}o{D}: Towards benchmarking consistency
  for task-oriented dialogue system}.
\newblock In \emph{Proceedings of the 2021 Conference on Empirical Methods in
  Natural Language Processing}, pages 2357--2367, Online and Punta Cana,
  Dominican Republic. Association for Computational Linguistics.

\bibitem[{Raffel et~al.(2020)Raffel, Shazeer, Roberts, Lee, Narang, Matena,
  Zhou, Li, and Liu}]{2020t5}
Colin Raffel, Noam Shazeer, Adam Roberts, Katherine Lee, Sharan Narang, Michael
  Matena, Yanqi Zhou, Wei Li, and Peter~J. Liu. 2020.
\newblock \href {http://jmlr.org/papers/v21/20-074.html} {Exploring the limits
  of transfer learning with a unified text-to-text transformer}.
\newblock \emph{Journal of Machine Learning Research}, 21(140):1--67.

\bibitem[{Rashkin et~al.(2021{\natexlab{a}})Rashkin, Nikolaev, Lamm, Collins,
  Das, Petrov, Tomar, Turc, and Reitter}]{rashkin2021measuring}
Hannah Rashkin, Vitaly Nikolaev, Matthew Lamm, Michael Collins, Dipanjan Das,
  Slav Petrov, Gaurav~Singh Tomar, Iulia Turc, and David Reitter.
  2021{\natexlab{a}}.
\newblock Measuring attribution in natural language generation models.
\newblock \emph{arXiv preprint arXiv:2112.12870}.

\bibitem[{Rashkin et~al.(2021{\natexlab{b}})Rashkin, Reitter, Tomar, and
  Das}]{rashkin-etal-2021-increasing}
Hannah Rashkin, David Reitter, Gaurav~Singh Tomar, and Dipanjan Das.
  2021{\natexlab{b}}.
\newblock \href {https://doi.org/10.18653/v1/2021.acl-long.58} {Increasing
  faithfulness in knowledge-grounded dialogue with controllable features}.
\newblock In \emph{Proceedings of the 59th Annual Meeting of the Association
  for Computational Linguistics and the 11th International Joint Conference on
  Natural Language Processing (Volume 1: Long Papers)}, pages 704--718, Online.
  Association for Computational Linguistics.

\bibitem[{Reiter and Thomson(2020)}]{reiter-thomson-2020-shared}
Ehud Reiter and Craig Thomson. 2020.
\newblock \href {https://aclanthology.org/2020.inlg-1.28} {Shared task on
  evaluating accuracy}.
\newblock In \emph{Proceedings of the 13th International Conference on Natural
  Language Generation}, pages 227--231, Dublin, Ireland. Association for
  Computational Linguistics.

\bibitem[{Rohrbach et~al.(2018)Rohrbach, Hendricks, Burns, Darrell, and
  Saenko}]{rohrbach2018object}
Anna Rohrbach, Lisa~Anne Hendricks, Kaylee Burns, Trevor Darrell, and Kate
  Saenko. 2018.
\newblock Object hallucination in image captioning.
\newblock In \emph{Proceedings of the 2018 Conference on Empirical Methods in
  Natural Language Processing}, pages 4035--4045.

\bibitem[{Schuster et~al.(2021)Schuster, Fisch, and
  Barzilay}]{schuster-etal-2021-get}
Tal Schuster, Adam Fisch, and Regina Barzilay. 2021.
\newblock \href {https://doi.org/10.18653/v1/2021.naacl-main.52} {Get your
  vitamin {C}! robust fact verification with contrastive evidence}.
\newblock In \emph{Proceedings of the 2021 Conference of the North American
  Chapter of the Association for Computational Linguistics: Human Language
  Technologies}, pages 624--643, Online. Association for Computational
  Linguistics.

\bibitem[{Scialom et~al.(2021)Scialom, Dray, Lamprier, Piwowarski, Staiano,
  Wang, and Gallinari}]{scialom-etal-2021-questeval}
Thomas Scialom, Paul-Alexis Dray, Sylvain Lamprier, Benjamin Piwowarski, Jacopo
  Staiano, Alex Wang, and Patrick Gallinari. 2021.
\newblock \href {https://aclanthology.org/2021.emnlp-main.529} {{Q}uest{E}val:
  Summarization asks for fact-based evaluation}.
\newblock In \emph{Proceedings of the 2021 Conference on Empirical Methods in
  Natural Language Processing}, pages 6594--6604, Online and Punta Cana,
  Dominican Republic. Association for Computational Linguistics.

\bibitem[{Scialom and Hill(2021)}]{scialom2021beametrics}
Thomas Scialom and Felix Hill. 2021.
\newblock Beametrics: A benchmark for language generation evaluation
  evaluation.
\newblock \emph{arXiv preprint arXiv:2110.09147}.

\bibitem[{Sellam et~al.(2020{\natexlab{a}})Sellam, Das, and
  Parikh}]{sellam-etal-2020-bleurt}
Thibault Sellam, Dipanjan Das, and Ankur Parikh. 2020{\natexlab{a}}.
\newblock \href {https://doi.org/10.18653/v1/2020.acl-main.704} {{BLEURT}:
  Learning robust metrics for text generation}.
\newblock In \emph{Proceedings of the 58th Annual Meeting of the Association
  for Computational Linguistics}, pages 7881--7892, Online. Association for
  Computational Linguistics.

\bibitem[{Sellam et~al.(2020{\natexlab{b}})Sellam, Pu, Chung, Gehrmann, Tan,
  Freitag, Das, and Parikh}]{sellam-etal-2020-learning}
Thibault Sellam, Amy Pu, Hyung~Won Chung, Sebastian Gehrmann, Qijun Tan, Markus
  Freitag, Dipanjan Das, and Ankur Parikh. 2020{\natexlab{b}}.
\newblock \href {https://aclanthology.org/2020.wmt-1.102} {Learning to evaluate
  translation beyond {E}nglish: {BLEURT} submissions to the {WMT} metrics 2020
  shared task}.
\newblock In \emph{Proceedings of the Fifth Conference on Machine Translation},
  pages 921--927, Online. Association for Computational Linguistics.

\bibitem[{Shaham et~al.(2022)Shaham, Segal, Ivgi, Efrat, Yoran, Haviv, Gupta,
  Xiong, Geva, Berant, and Levy}]{shaham2022scrolls}
Uri Shaham, Elad Segal, Maor Ivgi, Avia Efrat, Ori Yoran, Adi Haviv, Ankit
  Gupta, Wenhan Xiong, Mor Geva, Jonathan Berant, and Omer Levy. 2022.
\newblock \href {http://arxiv.org/abs/2201.03533} {Scrolls: Standardized
  comparison over long language sequences}.

\bibitem[{Thorne et~al.(2018)Thorne, Vlachos, Cocarascu, Christodoulopoulos,
  and Mittal}]{thorne-etal-2018-fact}
James Thorne, Andreas Vlachos, Oana Cocarascu, Christos Christodoulopoulos, and
  Arpit Mittal. 2018.
\newblock \href {https://doi.org/10.18653/v1/W18-5501} {The fact extraction and
  {VER}ification ({FEVER}) shared task}.
\newblock In \emph{Proceedings of the First Workshop on Fact Extraction and
  {VER}ification ({FEVER})}, pages 1--9, Brussels, Belgium. Association for
  Computational Linguistics.

\bibitem[{Wang et~al.(2020)Wang, Cho, and Lewis}]{wang-etal-2020-asking}
Alex Wang, Kyunghyun Cho, and Mike Lewis. 2020.
\newblock \href {https://doi.org/10.18653/v1/2020.acl-main.450} {Asking and
  answering questions to evaluate the factual consistency of summaries}.
\newblock In \emph{Proceedings of the 58th Annual Meeting of the Association
  for Computational Linguistics}, pages 5008--5020, Online. Association for
  Computational Linguistics.

\bibitem[{Wang et~al.(2018)Wang, Singh, Michael, Hill, Levy, and
  Bowman}]{wang-etal-2018-glue}
Alex Wang, Amanpreet Singh, Julian Michael, Felix Hill, Omer Levy, and Samuel
  Bowman. 2018.
\newblock \href {https://doi.org/10.18653/v1/W18-5446} {{GLUE}: A multi-task
  benchmark and analysis platform for natural language understanding}.
\newblock In \emph{Proceedings of the 2018 {EMNLP} Workshop {B}lackbox{NLP}:
  Analyzing and Interpreting Neural Networks for {NLP}}, pages 353--355,
  Brussels, Belgium. Association for Computational Linguistics.

\bibitem[{Welleck et~al.(2019)Welleck, Weston, Szlam, and
  Cho}]{welleck-etal-2019-dialogue}
Sean Welleck, Jason Weston, Arthur Szlam, and Kyunghyun Cho. 2019.
\newblock \href {https://doi.org/10.18653/v1/P19-1363} {Dialogue natural
  language inference}.
\newblock In \emph{Proceedings of the 57th Annual Meeting of the Association
  for Computational Linguistics}, pages 3731--3741, Florence, Italy.
  Association for Computational Linguistics.

\bibitem[{Xie et~al.(2021)Xie, Sun, Deng, Li, and
  Ding}]{xie-etal-2021-factual-consistency}
Yuexiang Xie, Fei Sun, Yang Deng, Yaliang Li, and Bolin Ding. 2021.
\newblock \href {https://aclanthology.org/2021.findings-emnlp.10} {Factual
  consistency evaluation for text summarization via counterfactual estimation}.
\newblock In \emph{Findings of the Association for Computational Linguistics:
  EMNLP 2021}, pages 100--110, Punta Cana, Dominican Republic. Association for
  Computational Linguistics.

\bibitem[{Yeh et~al.(2021)Yeh, Eskenazi, and
  Mehri}]{yeh-etal-2021-comprehensive}
Yi-Ting Yeh, Maxine Eskenazi, and Shikib Mehri. 2021.
\newblock \href {https://aclanthology.org/2021.eancs-1.3} {A comprehensive
  assessment of dialog evaluation metrics}.
\newblock In \emph{The First Workshop on Evaluations and Assessments of Neural
  Conversation Systems}, pages 15--33, Online. Association for Computational
  Linguistics.

\bibitem[{Yin et~al.(2021)Yin, Radev, and Xiong}]{yin-etal-2021-docnli}
Wenpeng Yin, Dragomir Radev, and Caiming Xiong. 2021.
\newblock \href {https://doi.org/10.18653/v1/2021.findings-acl.435}
  {{D}oc{NLI}: A large-scale dataset for document-level natural language
  inference}.
\newblock In \emph{Findings of the Association for Computational Linguistics:
  ACL-IJCNLP 2021}, pages 4913--4922, Online. Association for Computational
  Linguistics.

\bibitem[{Yuan et~al.(2021)Yuan, Neubig, and Liu}]{yuan2021bartscore}
Weizhe Yuan, Graham Neubig, and Pengfei Liu. 2021.
\newblock \href {https://openreview.net/forum?id=5Ya8PbvpZ9} {{BARTS}core:
  Evaluating generated text as text generation}.
\newblock In \emph{Advances in Neural Information Processing Systems}.

\bibitem[{Zhang et~al.(2020)Zhang, Kishore, Wu, Weinberger, and
  Artzi}]{bert-score}
Tianyi Zhang, Varsha Kishore, Felix Wu, Kilian~Q. Weinberger, and Yoav Artzi.
  2020.
\newblock \href {https://openreview.net/forum?id=SkeHuCVFDr} {Bertscore:
  Evaluating text generation with bert}.
\newblock In \emph{International Conference on Learning Representations}.

\bibitem[{Zhang et~al.(2019)Zhang, Baldridge, and He}]{zhang-etal-2019-paws}
Yuan Zhang, Jason Baldridge, and Luheng He. 2019.
\newblock \href {https://doi.org/10.18653/v1/N19-1131} {{PAWS}: Paraphrase
  adversaries from word scrambling}.
\newblock In \emph{Proceedings of the 2019 Conference of the North {A}merican
  Chapter of the Association for Computational Linguistics: Human Language
  Technologies, Volume 1 (Long and Short Papers)}, pages 1298--1308,
  Minneapolis, Minnesota. Association for Computational Linguistics.

\bibitem[{Zhao et~al.(2020)Zhao, Cohen, and Webber}]{zhao2020reducing}
Zheng Zhao, Shay~B Cohen, and Bonnie Webber. 2020.
\newblock Reducing quantity hallucinations in abstractive summarization.
\newblock \emph{arXiv preprint arXiv:2009.13312}.

\bibitem[{Zhou et~al.(2021)Zhou, Neubig, Gu, Diab, Guzm{\'a}n, Zettlemoyer, and
  Ghazvininejad}]{zhou-etal-2021-detecting}
Chunting Zhou, Graham Neubig, Jiatao Gu, Mona Diab, Francisco Guzm{\'a}n, Luke
  Zettlemoyer, and Marjan Ghazvininejad. 2021.
\newblock \href {https://doi.org/10.18653/v1/2021.findings-acl.120} {Detecting
  hallucinated content in conditional neural sequence generation}.
\newblock In \emph{Findings of the Association for Computational Linguistics:
  ACL-IJCNLP 2021}, pages 1393--1404, Online. Association for Computational
  Linguistics.

\end{thebibliography}

\clearpage

\appendix

\section{Additional Data Statistics}\label{sec:data_statistic}
Tables \ref{tab:grounding_length} and \ref{tab:generated_length} presents statistics regarding the length of the grounding text and the generated text for TRUE's datasets, respectively.

\begin{table}[ht!]
\begin{center}
\begin{small}
\scalebox{0.7}{
\begin{tabular}{|l|l|l|l|l|}
\hline
\textbf{Dataset} & \textbf{Min len.} & \textbf{Max len.} & \textbf{Median len.} & \textbf{Avg len.} \\ \hline
FRANK      & 102 & 1005  & 550 & 548 \\ \hline
SummEval        & 100 & 540   & 367 & 359 \\ \hline
MNBM         & 8   & 10315 & 287 & 383 \\ \hline
QAGS-CNNDM     & 73  & 360   & 325 & 318 \\ \hline
QAGS-XSUM      & 218 & 520   & 339 & 351 \\ \hline
BEGIN      & 7   & 64    & 23  & 23  \\ \hline
$Q^{2}$              & 6   & 71    & 21  & 23  \\ \hline
DialFact & 4   & 174   & 22  & 26  \\ \hline
PAWS & 5   & 37    & 21.0  & 21  \\ \hline
FEVER       & 8   & 286   & 44  & 59  \\ \hline
VitaminC       & 1   & 265   & 26  & 28  \\ \hline
\end{tabular}
}
\end{small}
\end{center}
\caption{Grounding length statistics for TRUE.}
\label{tab:grounding_length}
\end{table}

\begin{table}[ht!]
\begin{center}
\begin{small}
\scalebox{0.7}{
\begin{tabular}{|l|l|l|l|l|}
\hline
\textbf{Dataset} & \textbf{Min len.} & \textbf{Max len.} & \textbf{Median len.} & \textbf{Avg len.} \\ \hline
FRANK      & 2 & 126 & 40 & 41 \\ \hline
SummEval        & 5 & 133 & 61 & 63 \\ \hline
MNBM          & 2 & 52  & 19 & 19 \\ \hline
QAGS-CNNDM     & 23 & 85  & 47 & 49 \\ \hline
QAGS-XSUM      & 9 & 31  & 18 & 18 \\ \hline
BEGIN     & 5 & 40  & 13 & 14 \\ \hline
$Q^{2}$     & 7 & 44  & 15 & 16 \\ \hline
DialFact & 4 & 69  & 16 & 17 \\ \hline
PAWS & 5 & 37  & 21 & 21 \\ \hline
FEVER     & 2 & 36  & 8  & 8  \\ \hline
VitaminC       & 1   & 103   & 12  & 13  \\ \hline
\end{tabular}
}
\end{small}
\end{center}
\caption{Generated text length statistics for TRUE.}
\label{tab:generated_length}
\end{table}

\begin{table*}[!ht]
\begin{center}
\begin{small}
\begin{tabular}{|l|l|l|l|l|l|l|}
\hline
\textbf{Dataset} & \textbf{Pos ROUGE\_L} & \textbf{Neg ROUGE\_L} & \textbf{ROUGE\_L diff} & \textbf{Pos F1} & \textbf{Neg F1} & \textbf{F1 diff} \\ \hline
FRANK           & 0.105 & 0.060 & 0.045 & 0.165 & 0.103 & 0.062 \\ \hline
SummEval        & 0.181 & 0.141 & 0.041 & 0.282 & 0.244 & 0.038 \\ \hline
MNBM            & 0.044 & 0.047 & 0.003 & 0.079 & 0.084 & 0.006 \\ \hline
QAGS-CNNDM      & 0.215 & 0.170 & 0.045 & 0.281 & 0.249 & 0.031 \\ \hline
QAGS-XSUM       & 0.051 & 0.050 & 0.002 & 0.082 & 0.080 & 0.002 \\ \hline
BEGIN           & 0.465 & 0.159 & 0.306 & 0.553 & 0.207 & 0.346 \\ \hline
$Q^{2}$         & 0.228 & 0.169 & 0.059 & 0.368 & 0.264 & 0.104 \\ \hline
DialFact        & 0.302 & 0.200 & 0.102 & 0.394 & 0.249 & 0.144 \\ \hline
PAWS            & 0.832 & 0.734 & 0.098 & 0.938 & 0.934 & 0.003 \\ \hline
FEVER     & 0.174 & 0.179 & 0.005 & 0.276 & 0.258 & 0.018 \\ \hline
VitaminC    & 0.314 & 0.270 & 0.044 & 0.362 & 0.290 & 0.072 \\ \hline
\end{tabular}
\end{small}
\end{center}
\caption{Average overlap between the generated text and the grounding, measured using ROUGE-L and simple F1 token-overlap, taking the grounding to be the reference text. The ``Pos'' columns contain the statistics for the grounded text, while the ``Neg'' columns contain the statistics for the ungrounded text.}
\label{tab:overlap_stat}
\end{table*}

\section{Implementation Details}
\label{sec:implementation_details}
We train all models using the t5x library.\footnote{https://github.com/google-research/t5x}

\paragraph{QG-QA} For our reimplementation of $Q^2$ \cite{honovich-etal-2021-q2} we use T5-11B as the pretrained model for QG, QA and NLI, while \citet{honovich-etal-2021-q2} used T5-Base, ALBERT \cite{lan2019albert}, and RoBERTa \cite{liu2019roberta} for the QG, QA and NLI models, respectively. We use a maximum length of 2048 tokens for the input. We set the F1 token overlap threshold to 0.54 by tuning it on a held-out dataset. We use beam search with a beam size of 4 to generate multiple questions, and use the first question that passes the validation threshold.

\paragraph{NLI} We fine-tune a T5-11B model on ANLI~\cite{nie-etal-2020-adversarial} for $25K$ steps with a learning rate of $10^{-4}$ and a batch size of 32. During inference we use a maximum input length of 2048 tokens.


\section{ROC Curves}

\begin{figure*}[ht!]
    \centering
    \includegraphics[width=.32\textwidth]{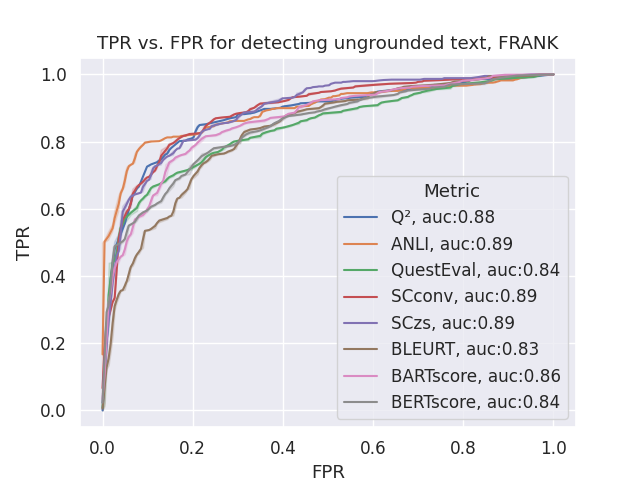}
    \includegraphics[width=.32\textwidth]{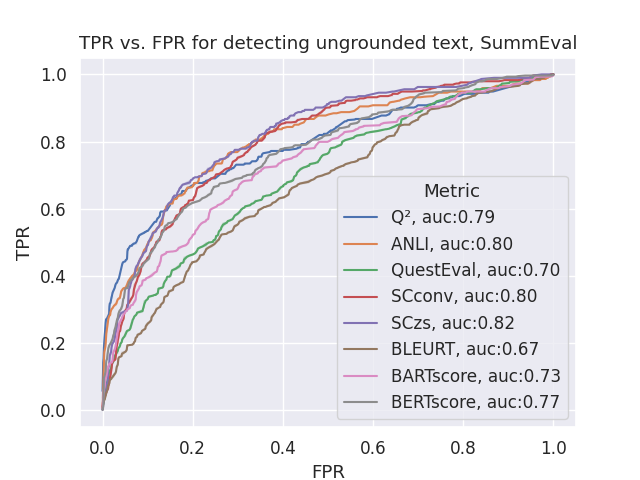}
    \includegraphics[width=.32\textwidth]{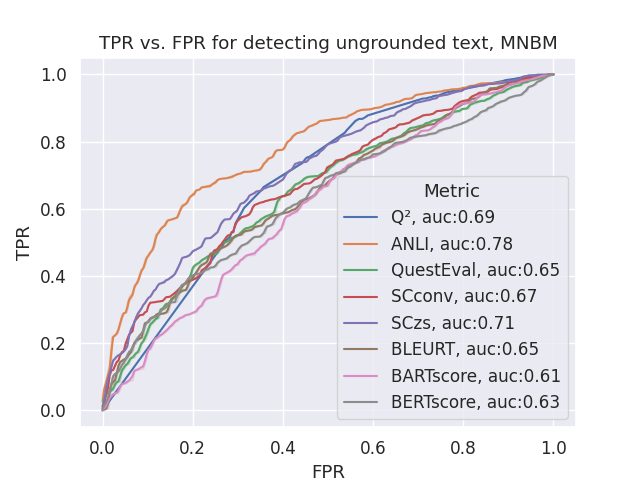}
    \\
    \includegraphics[width=.32\textwidth]{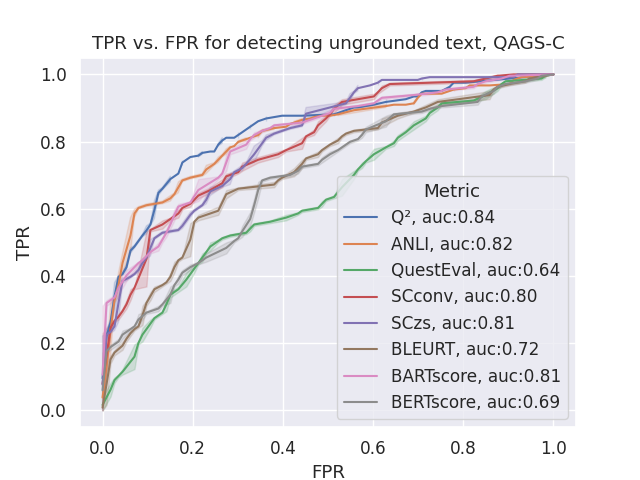}
    \includegraphics[width=.32\textwidth]{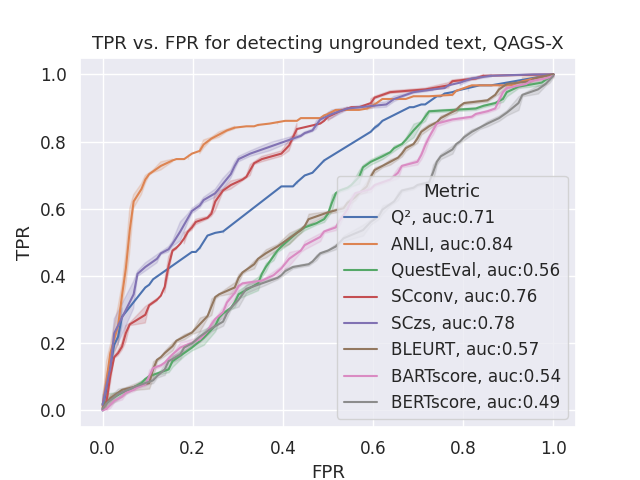}
    \includegraphics[width=.32\textwidth]{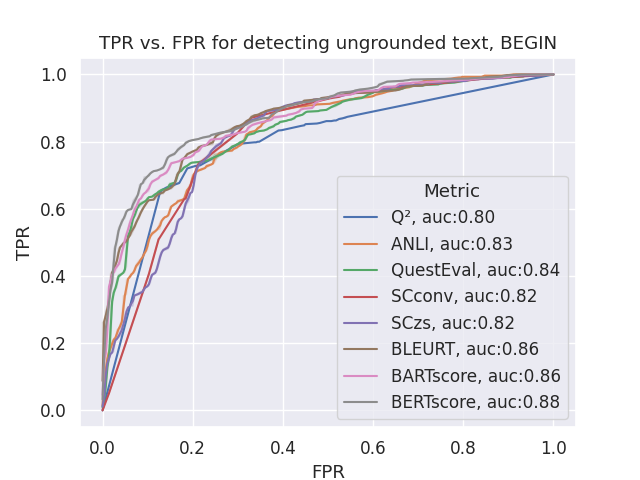}
    \\
    \includegraphics[width=.32\textwidth]{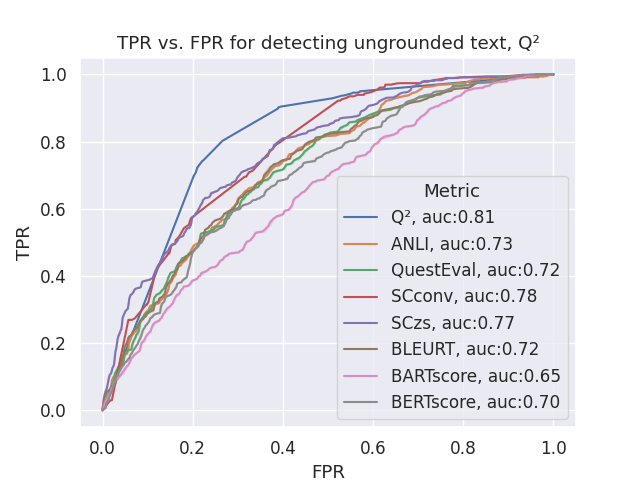}
    \includegraphics[width=.32\textwidth]{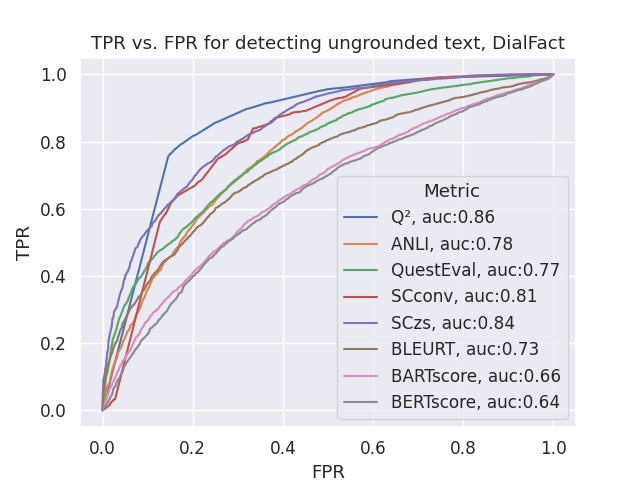}
    \includegraphics[width=.32\textwidth]{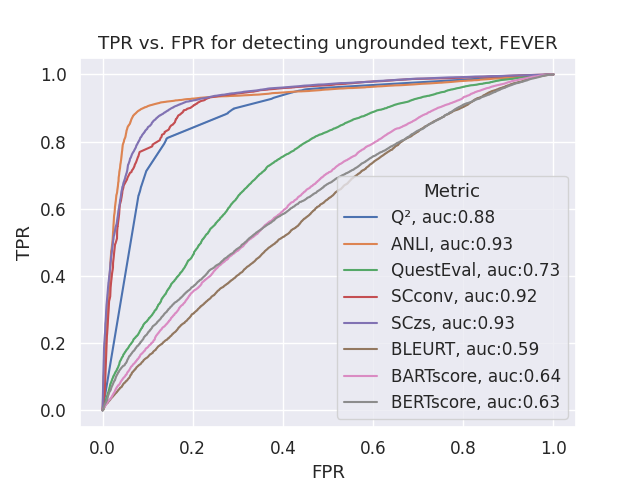}
    \\
    \includegraphics[width=.32\textwidth]{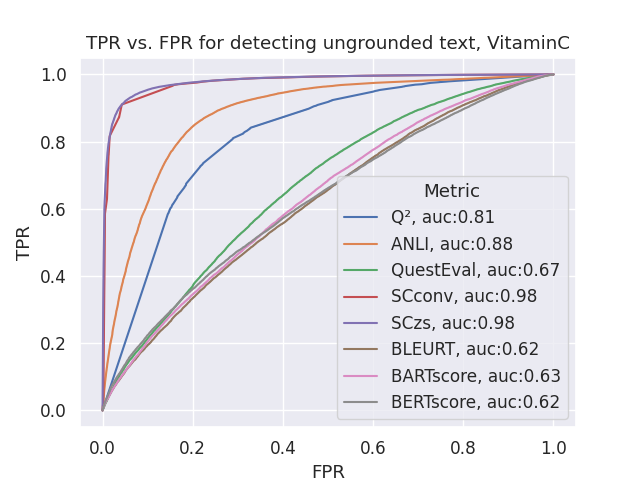}
    \includegraphics[width=.32\textwidth]{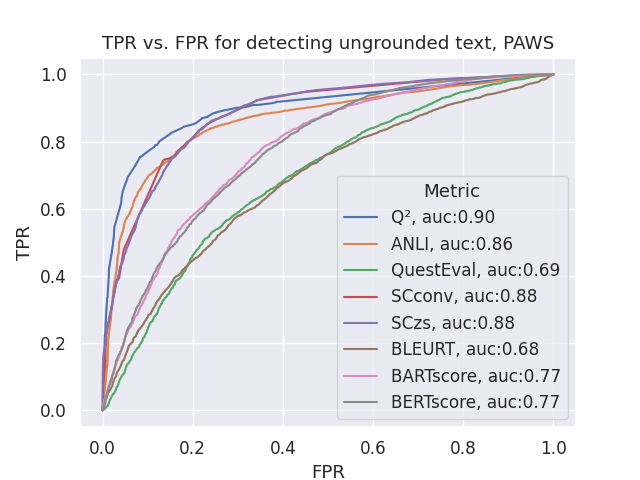}
    \caption{ROC curves for the best performing methods.}
    \label{fig:roc_curves}
\end{figure*}

Figure \ref{fig:roc_curves} presents the ROC curves for the different datasets studied in TRUE, using the best-performing metrics.

\begin{table*}[!t]
\begin{center}
\begin{small}
\scalebox{0.82}{
\begin{tabular}{|l|l|l|l|l|l|l|l|}
\hline
                     & \multicolumn{1}{l|}{\textbf{ANLI}+$\mathbf{Q^2}$} & \multicolumn{1}{l|}{\textbf{ANLI}+\textbf{\summaczs{}}} & \multicolumn{1}{l|}{$\mathbf{Q^2}$+\textbf{\summaczs{}}} & \multicolumn{1}{l|}{\textbf{CTC}} & \multicolumn{1}{l|}{\textbf{\summacconv{}}} & \multicolumn{1}{l|}{\textbf{ROUGE-L}} & \multicolumn{1}{l|}{\textbf{BLEU4}}
\\ \hline\hline
\textbf{FRANK}   & 89.6 & 91.1 & 90.4 & 87.5 & 88.9             & 80.1                       & 78.0                  \\ \hline
\textbf{SummEval}& 80.7 & 83.0 & 82.0 & 76.0 & 79.8             & 68.8                       & 60.2                    \\ \hline
\textbf{MNBM}    & 75.6 & 77.1 & 74.6 & 72.3 & 67.2             & 47.5                       & 49.3                   \\ \hline
\textbf{QAGS-C}  & 86.0 & 84.7& 86.4 & 73.4 & 79.6             & 67.1                       & 63.9                   \\ \hline
\textbf{QAGS-X}  & 81.8 & 85.1 & 79.3 & 73.1 & 76.1             & 52.9                       & 48.6                    \\ \hline
\textbf{BEGIN}   & 85.7 & 82.1 & 85.7 & 77.9 & 81.6              & 86.4                       & 84.6                \\ \hline
$\mathbf{Q^2}$   & 83.0 & 76.9 & 83.9 & 85.3 & 77.5             & 66.8                       & 64.3                    \\ \hline
\textbf{DialFact}& 89.4 & 84.5 & 90.2 & 83.5 & 81.2             & 71.2                       & 72.5                    \\ \hline
\textbf{PAWS}    & 90.5 & 89.7 & 91.4 & 86.0 & 88.2             & 82.2                       & 77.3                   \\ \hline\hline
\textbf{FEVER}   & 94.0 & \st{94.6} & \st{93.9} & 84.8 & \st{86.7}         & 49.9                       & 51.1                   \\ \hline
\textbf{VitaminC}& 90.3  & \st{96.4} & \st{96.5}  & 84.9 & \st{97.5}       & 59.9                       & 59.6                 \\ \hline\hline
\textbf{Avg. $_{\text{w/o VitC, FEVER}}$}    
                 &84.7 & 83.8 & 84.9 & 79.4 & 80.0             & 69.2                       & 66.5               \\ \hline
\end{tabular}
}
\end{small}
\end{center}
\caption{ROC AUC results for metrics that were not reported in Table \ref{tab:auc}.}  
\label{tab:auc_weak_metrics}
\end{table*}


\begin{table*}[!ht]
\begin{center}
\begin{small}
\scalebox{0.82}{
\begin{tabular}{|l||l||l|l|l|l|l|l|l|l|l|}
\hline
                      & \multicolumn{1}{l||}{\textbf{Ensemble}} & \multicolumn{1}{l|}{$\mathbf{Q^2}$} & \multicolumn{1}{l|}{\textbf{ANLI}}  & \multicolumn{1}{l|}{\textbf{\summaczs{}}} & \multicolumn{1}{l|}{\textbf{BLEURT}} & \multicolumn{1}{l|}{\textbf{QuestEval}} &  \multicolumn{1}{l|}{\textbf{FactCC}} & \multicolumn{1}{l|}{\textbf{BART$_{\text{score}}$}} & \multicolumn{1}{l|}{\textbf{BERT$_{\text{score}}$}}
\\ \hline\hline
\textbf{FRANK}   & 90.8             & 87.8    & \textbf{89.2}                       & 88.6          & 83.2 & 86.4    & 73.9 & 88.3 & 86.0          \\ \hline
\textbf{BEGIN}   & 85.9             & 78.0             & 82.8                       & 84.2            & 82.2 & 81.4  & 65.0 & 83.7 & \textbf{86.0} \\ \hline
\textbf{DialFact}& 88.6             & \textbf{85.0}             & 75.9                       & 82.1          & 72.2 & 76.3  & 55.1 & 65.5 & 64.3          \\ \hline
\textbf{PAWS}    & 92.4             & \textbf{90.1}             & 87.3                       & 89.7          & 67.1 & 70.1  & 65.1   & 77.3 & 76.4          \\ \hline
\textbf{VitaminC} & 96.7             & 83.4            & \textbf{89.6}                       & \st{98.4}     & 63.0 & 67.8  & 56.8 & 64.1 & 63.5          \\ \hline\hline
\textbf{Avg. $_{\text{w/o VitC}}$}    
                 & 89.4              & 85.2    & 83.8                    & \textbf{86.2}    & 76.2 & 78.5 &  64.8 & 78.7 & 78.2          \\ \hline
\end{tabular}
}
\end{small}
\end{center}
\caption{ROC AUC results for the different metrics on the TRUE test set. We exclude VitaminC from the average calculation as \summaczs{} was trained on VitaminC. The highest score in each row (excluding the Ensemble) is in bold and the aforementioned SC results are in strikethrough.}
\label{tab:auc_test}
\end{table*}

\begin{table*}[!ht]
\begin{center}
\begin{small}
\scalebox{0.88}{
\begin{tabular}{|l||l||l|l|l|l|l|l|l|l|l|l|}
\hline
                      & \multicolumn{1}{l||}{\textbf{Ensemble}} & \multicolumn{1}{l|}{$\mathbf{Q^2}_{\textbf{metric}}$} & \multicolumn{1}{l|}{\textbf{ANLI}}  & \multicolumn{1}{l|}{\textbf{\summaczs{}}} & \multicolumn{1}{l|}{\textbf{F1}} & \multicolumn{1}{l|}{\textbf{BLEURT}} & \multicolumn{1}{l|}{\textbf{QuestEval}} &  \multicolumn{1}{l|}{\textbf{FactCC}} & \multicolumn{1}{l|}{\textbf{BART$_{\text{score}}$}} & \multicolumn{1}{l|}{\textbf{BERT$_{\text{score}}$}}
\\ \hline\hline
\textbf{FRANK}   & 83.2 & 82.9              & \textbf{83.5}    & 80.9                       & 67.5          & 76.0 & 74.2    & 73.9 & 80.2 & 77.0          \\ \hline
\textbf{SummEval}& 79.4 & \textbf{77.3}     & 72.9             & 75.0                       & 64.2          & 66.4 & 67.0  & 65.3 & 68.9 & 73.7           \\ \hline
\textbf{MNBM}    & 69.6 & 66.5              & \textbf{66.7}  & 64.5                       & 40.4          & 53.7 & 66.1  & 62.9 & 61.0 & 60.1          \\ \hline
\textbf{QAGS-C}  & 81.3 & \textbf{78.3}     & 75.3             & 72.8                       & 62.6          & 68.9 & 62.6  & 70.6 & 74.5 & 66.8          \\ \hline
\textbf{QAGS-X}  & 77.4 & 63.6              & \textbf{80.3}    & 72.8                       & 51.0          & 56.5 & 56.9  & 65.3 & 54.4 & 52.3          \\ \hline
\textbf{BEGIN}   & 77.9 & 73.0              & 76.0             & 79.7                       & 80.0           & 78.1 & 75.5  & 66.6 & 77.4 & \textbf{80.5} \\ \hline
$\mathbf{Q^2}_{\textbf{dataset}}$   & 76.8 & \textbf{76.1}    & 66.6             & 71.3                       & 61.4          & 67.0 & 65.8  & 60.5 & 59.7 & 66.1          \\ \hline
\textbf{DialFact}& 83.4 & \textbf{80.5}   & 70.9            & 75.8                       & 66.4          & 67.0 & 69.9  & 53.4 & 60.8 & 61.2          \\ \hline
\textbf{PAWS}    & 83.4 & \textbf{83.5}   & 80.5             & 81.5                       & 50.2          & 64.2 & 64.3  & 60.2   & 72.2 & 71.1          \\ \hline\hline
\textbf{FEVER}   & 90.4 & 82.7              & \textbf{90.2}  & \st{87.6}                  & 52.3          & 57.7 & 69.4  & 57.2 & 60.5 & 58.7          \\ \hline
\textbf{VitaminC}& 90.8 & 76.0              & \textbf{82.3}  & \st{93.5}                  & 58.0          & 58.0 & 62.2  & 55.0 & 59.1 & 58.6          \\ \hline\hline
\textbf{Avg. $_{\text{w/o VitC, FEVER}}$}    
                 & 79.2 & \textbf{75.7}              & 74.7    & 74.9                       & 60.4          & 66.4 & 66.9 &  64.3 & 67.6 & 67.6          \\ \hline
\end{tabular}
}
\end{small}
\end{center}
\caption{Accuracy results for the different metrics on the TRUE development set. Note that thresholds were tuned on the development set itself. We exclude VitaminC and FEVER from the average calculation as \summaczs{} was trained on VitaminC that includes examples from FEVER. The highest score in each row (excluding the Ensemble) is in bold and the aforementioned SC results are in strikethrough.}
\label{tab:acc_dev}
\end{table*}

\begin{table*}[!ht]
\begin{center}
\begin{small}
\scalebox{0.82}{
\begin{tabular}{|l||l||l|l|l|l|l|l|l|l|l|}
\hline
                      & \multicolumn{1}{l||}{\textbf{Ensemble}} & \multicolumn{1}{l|}{$\mathbf{Q^2}$} & \multicolumn{1}{l|}{\textbf{ANLI}}  & \multicolumn{1}{l|}{\textbf{\summaczs{}}} & \multicolumn{1}{l|}{\textbf{BLEURT}} & \multicolumn{1}{l|}{\textbf{QuestEval}} &  \multicolumn{1}{l|}{\textbf{FactCC}} & \multicolumn{1}{l|}{\textbf{BART$_{\text{score}}$}} & \multicolumn{1}{l|}{\textbf{BERT$_{\text{score}}$}}
\\ \hline\hline
\textbf{FRANK}   & 83.0             & 81.5    & \textbf{82.0}                       & 79.0          & 76.6 & 73.0    & 72.1 & 80.7 & 75.6          \\ \hline
\textbf{BEGIN}   & 76.8             & 74.1             & 76.8                       & \textbf{78.9}            & 74.3 & 73.4  & 62.09 & 74.8 & 78.1 \\ \hline
\textbf{DialFact}& 80.9             & \textbf{78.1}     & 68.4                       & 74.2          & 67.1 & 69.0  & 52.5 & 58.6 & 60.2          \\ \hline
\textbf{PAWS}    & 84.8             & \textbf{84.1}     & 82.1                       & 82.3          & 62.9 & 64.8  & 60.7   & 70.9 & 69.8          \\ \hline
\textbf{VitaminC} & 92.1             & 77.5            & \textbf{83.9}                       & \st{94.2}     & 59.0 & 63.3  & 55.5 & 59.8 & 58.0          \\ \hline\hline
\textbf{Avg. $_{\text{w/o VitC}}$}    
                 & 81.4              & \textbf{79.4}    & 77.3                    & 78.6    & 70.2 & 70.0 &  62.1 & 71.3 & 70.9          \\ \hline
\end{tabular}
}
\end{small}
\end{center}
\caption{Accuracy results for the different metrics on the TRUE test set. Thresholds were tuned on the corresponding development sets. We exclude VitaminC from the average calculation as \summaczs{} was trained on VitaminC. The highest score in each row (excluding the Ensemble) is in bold and the aforementioned SC results are in strikethrough.}
\label{tab:accuracy_test}
\end{table*}
\end{document}